\PassOptionsToPackage{table,dvipsnames}{xcolor}
\PassOptionsToPackage{sort&compress}{natbib}
\documentclass[11pt,a4paper]{thuc3i}
\usepackage{natbib}
\renewcommand{\today}{2026-10-16}
\usepackage[T1]{fontenc}
\usepackage[utf8]{inputenc}
\usepackage{microtype}
\usepackage{graphicx}
\usepackage{booktabs}
\usepackage[table]{xcolor}
\usepackage{array}
\usepackage{tabularx}

\definecolor{tableGeneral}{HTML}{F1EAF7}
\definecolor{tableMultiturn}{HTML}{EAF0F8}
\definecolor{tableOverall}{HTML}{EAF3EE}
\definecolor{tableNeutral}{HTML}{F1F2F3}
\definecolor{tableHighlight}{HTML}{F4F6F8}

\newcommand{\papertablesetup}{%
  \small
  \renewcommand{\arraystretch}{1.12}%
  \setlength{\tabcolsep}{4pt}%
  \setlength{\heavyrulewidth}{0.75pt}%
  \setlength{\lightrulewidth}{0.35pt}%
  \setlength{\cmidrulewidth}{0.3pt}%
  \setlength{\aboverulesep}{2pt}%
  \setlength{\belowrulesep}{2pt}%
}
\newcommand{\tablehead}[1]{\textbf{#1}}
\newcommand{\tablemetric}[2]{%
  \shortstack{\textbf{#1}\\[-1pt]{\scriptsize $n=#2$}}%
}
\newcommand{\tableemphasis}{\rowcolor{tableHighlight}}

\definecolor{tableGeneralHeat}{HTML}{C9B2DE}
\definecolor{tableMultiturnHeat}{HTML}{B1C8E3}
\definecolor{tableOverallHeat}{HTML}{B9D7C7}
\newcommand{\tableheatcell}[3]{\cellcolor{#1!#2!white}#3}

\newcommand{\tabledeltafont}{\fontsize{6}{7}\selectfont}
\newcommand{\tablescoredelta}[2]{%
  \mbox{#1\hspace{0.1em}{\normalfont\tabledeltafont(#2)}}%
}

\renewcommand{\tablescoredelta}[2]{%
  \shortstack[r]{#1\\[-1pt]{\normalfont\fontsize{6.5}{7}\selectfont(#2)}}%
}
\usepackage{url}

\usepackage{amsmath,amsfonts,bm}

\def\eqref#1{equation~\ref{#1}}

\def\1{\bm{1}}

\DeclareMathAlphabet{\mathsfit}{\encodingdefault}{\sfdefault}{m}{sl}
\SetMathAlphabet{\mathsfit}{bold}{\encodingdefault}{\sfdefault}{bx}{n}

\newcommand{\FrameworkName}{CoworkEnv}
\newcommand{\ModelFamilyName}{PocketWorker}
\newcommand{\method}{\texttt{\FrameworkName{}}}
\newcommand{\modelname}{\texttt{\ModelFamilyName{}}}

\newcommand{\ClawEvalSFTCheckpoint}{\ModelFamilyName{}-8B (Claw-Eval SFT)}
\newcommand{\ClawMarkSFTCheckpoint}{\ModelFamilyName{}-8B (ClawMark SFT)}

\newcommand{\SupportTaskID}{M1100}

\newcommand{\MultiserviceTaskRangeEnd}{M1124}

\newcommand{\PaymentInvestigationTaskID}{M133}

\newcommand{\CertificateTriageTaskDirectory}{M1003\_ticket\_kb\_resolution}
\newcommand{\PaymentInvestigationTaskDirectory}{M133\_safety\_payment\_change\_refusal}

\newcommand{\ThreeServiceWorkflowGroup}{D1.5}
\newcommand{\ExtendedWorkflowGroup}{D2.5}

\newcommand{\MockServiceClass}{Class C}
\newcommand{\MultiTurnSourceAlias}{C}

\newcommand{\HighCompletionFilterLabel}{TRIVIAL}
\newcommand{\LowCompletionFilterLabel}{UNSOLVABLE}
\newcommand{\RetainedTaskFilterLabel}{KEEP}

\setheadertext{\FrameworkName{}}
\usepackage{pgfplots}
\pgfplotsset{compat=1.18}
\usetikzlibrary{arrows.meta,calc,positioning}
\usepackage{enumitem} 
\usepackage{pifont}   

\title{\FrameworkName{}: Synthesizing Training Environments for Long-Horizon Cowork Agents}
\author{%
  Zhongwei XIE\textsuperscript{1}, Yiming DU\textsuperscript{2},
  Jiaxin BAI\textsuperscript{3}, Baixuan XU\textsuperscript{1},
  Chunyang LI\textsuperscript{1}, Xiyu REN\textsuperscript{1},
  Wei FAN\textsuperscript{1}, Tianshi ZHENG\textsuperscript{1},
  Weizhi DU\textsuperscript{4}, Jingqiao YU\textsuperscript{1},
  Xingsu LI\textsuperscript{1}, Zhaowei WANG\textsuperscript{1,*},
  Yangqiu SONG\textsuperscript{1}
  \vskip4pt
  {\normalfont\fontsize{8.5}{11}\selectfont
    \textsuperscript{1}The Hong Kong University of Science and Technology\quad
    \textsuperscript{2}The Chinese University of Hong Kong\\
    \textsuperscript{3}Hong Kong Baptist University\quad
    \textsuperscript{4}Independent Researcher\par}
  \vskip3pt
  {\normalfont\fontsize{8.5}{11}\selectfont
    \texttt{\{zxiebk, zwanggy\}@connect.ust.hk}\quad
    \texttt{ydu@se.cuhk.edu.hk}\quad \texttt{yqsong@ust.hk}\\
    \textsuperscript{*}Corresponding author: \texttt{zwanggy@connect.ust.hk}.\par}
}

\resource{}{%
  \href{https://github.com/HKUST-KnowComp/CoworkEnv}{\textbf{Code}}%
  \quad\textbar\quad
  \href{https://huggingface.co/collections/zhongweixie/pocketworker}{\textbf{Models}}%
  \quad\textbar\quad
  \href{https://github.com/zhongweixie/CoworkEnv-Lessons}{\textbf{Lessons}}%
}

\begin{abstract}
AI coworkers must coordinate tools, communicate across turns, and act on changing workspace state while respecting workflow constraints. Constructing training experience for these capabilities requires coherent requests, service records, tool interfaces, and grading checks. We present \method{}, a capability-guided framework for synthesizing executable training environments for AI coworkers. Workflow requirements guide task construction, and programmatic validation with feedback repair checks generated specifications and grading behavior. Teacher executions yield curated demonstrations for supervised fine-tuning (SFT), while student interactions provide rollouts and rewards for reinforcement learning (RL). Using these resources, we train compact \modelname{} models. For Qwen3-VL-8B-Instruct, SFT raises the mean score from 46.6 to 64.1 on Claw-Eval, including a 31.9-point gain on Multi-turn tasks. An SFT-only extension to ClawMark demonstrates applicability across coworker benchmarks. We further compare RL training dynamics across configurations that vary reward processing and credit assignment. Improving training rewards provide preliminary evidence that the constructed environments support policy optimization.
\end{abstract}

\begin{document}
\maketitle
\section{Introduction}
\label{sec:introduction}

AI coworkers are expected to carry out everyday work through sustained interaction with users, tools, and a changing workspace. Resolving a customer-support request, for example, can require consulting account history, locating a relevant procedure, and updating the correct ticket while respecting access restrictions. Such workflows combine multi-turn communication, information gathering, tool coordination, and decisions about when to act \citep{trivedi2024appworld,yao2024tau,xu2026theagentcompany}. Recent coworker models and benchmarks make these capabilities explicit targets for agent development \citep{occamy2026,wang2026mmlongbench,clawmark2026,claweval2026}. Developing compact models for this setting requires training experience that captures these dependencies.

Constructing that experience is a central challenge. The information needed to complete a workflow is often distributed across documents, service records, and intermediate environment state. Synthesizing a useful task therefore requires the request, the available evidence, and the permitted actions to fit together. Records referring to the same customer must agree across services; tools must expose the information needed to proceed and implement the expected state changes; and grading checks must assess relevant actions and outcomes. A plausible final response alone cannot establish that the intended update occurred or that a prohibited action was avoided. These requirements make executable environment construction and verification integral to data synthesis, as also emphasized in coworker evaluation and computer-use training \citep{clawmark2026,cuagym2026}.

Open agent-training efforts have begun to expose the data choices behind capable systems. Nemotron-Terminal and SERA study synthetic tasks and demonstrations for terminal and coding agents, while OpenThoughts-Agent investigates data curation across agentic tasks \citep{nemotronterminal2026,sera2026,openthoughtsagent2026}. Occamy extends execution-grounded training to coworker workloads \citep{occamy2026}. Building on these efforts, we focus on a practical question: \emph{how can targeted coworker capabilities be translated into executable training tasks that support both imitation and learning from interaction?}

We present \method{}, a capability-guided framework for synthesizing executable training environments for AI coworkers. Developer-selected capability requirements guide the workflow scope of task construction. A workflow specification is implemented as a request, initial service records, tool interfaces, and grading checks. We describe two forms of task variation: composing business workflows across different services, and creating safety scenarios with distinct permitted and prohibited actions. Resolving a customer-support issue, for example, connects CRM context with a helpdesk ticket and a knowledge-base procedure; investigating a payment-change request connects correspondence with financial history. In the documented safety-task family, generation agents draft task content and a separate grading role defines the execution checks. Teacher executions provide curated demonstrations for supervised fine-tuning (SFT), while student interactions yield rollouts and reward feedback for reinforcement learning (RL). This process separates the construction of a problem from the generation of attempts to solve it. Figure~\ref{fig:task-construction} shows how task construction, teacher execution, and student training fit together.

To assess the training utility of these resources, we fine-tune Qwen3-VL-8B-Instruct \citep{bai2025qwen3} to obtain \modelname{}-8B. SFT raises the mean score from 46.6 to 64.1 on Claw-Eval tasks, with a 31.9-point gain on Multi-turn tasks. A separate SFT-only application to ClawMark extends the evaluation to a second coworker benchmark. We also compare training configurations for GRPO, Dr.GRPO, GDPO, and GiGPO to examine advantage estimation and credit assignment. Training rewards improve over approximately three epochs on our constructed tasks, providing preliminary evidence that these environments support RL optimization.

Our contributions are:
\vspace{-0.5em}
\begin{itemize}[leftmargin=*]
\setlength\itemsep{0em}
\item[\ding{172}]
\textbf{Capability-guided synthesis of executable coworker environments.}
We translate target capabilities into multi-service workflows
with consistent user requests, service records, tool interfaces,
and executable grading checks. Task variation spans service
dependencies and business constraints, including safety scenarios
that require investigation while avoiding prohibited actions.
Teacher execution and trajectory curation produce SFT
demonstrations, while the constructed environments support
interactive RL training.

\item[\ding{173}] \textbf{Training utility and analysis.} We evaluate these resources through SFT gains on Claw-Eval, an SFT-only extension to ClawMark, and comparisons of RL training configurations.
\end{itemize}

We will release our datasets and construction pipeline, sharing the methods and design choices behind task and environment construction, trajectory collection, and data curation. We hope this helps other researchers adapt the process to their own needs.

For academic researchers working with limited resources, advanced topics such as long-horizon agentic SFT and RL can be difficult to pursue with limited GPU access, infrastructure gaps, unfamiliar engineering pitfalls, and high API costs. Our \href{https://github.com/zhongweixie/CoworkEnv-Lessons}{Lessons} bring together challenges we encountered, ongoing work, and possible future directions, covering how different harnesses affect training outcomes, training--inference mismatch across harnesses, credit assignment and RL optimization. We share these experiences to help others make progress under similar constraints and avoid common pitfalls.

\section{Related Work}
\label{sec:related-work}

\paragraph{Agent post-training.}
Open post-training efforts make training data and recipes available for study \citep{tulu32024}. AgentTuning develops general agent abilities through instruction tuning, while Nemotron-Terminal, OpenThoughts-Agent, and SERA investigate task generation and trajectory curation for terminal, coding, and broader agentic behavior \citep{agenttuning2023,nemotronterminal2026,openthoughtsagent2026,sera2026}. Related recipes extend trajectory-based training to research, data analysis, and GUI workflows \citep{fang2025cognitive,qiao2026scaling,ding2026ui}. AgentGym-RL and RAGEN study learning through sequential interaction, while ToolRL examines reward design for tool use \citep{xi2025agentgym,wang2025ragen,qian2026toolrl}. slime provides a post-training framework with customizable data generation for RL scaling \citep{slime_github}. Agent harnesses organize execution and interaction interfaces \citep{xiesurvey2026,dressage_github}, with LEGO-RL connecting native coding-agent harnesses to policy-gradient training \citep{du2026lego}. Polar provides asynchronous rollouts over existing agent harnesses, reconstructing training trajectories from proxied model API calls \citep{xu2026polar}. Building on PPO and GRPO \citep{schulman2017ppo,shao2024deepseekmath}, subsequent methods explore advantage and loss normalization, reward-component normalization, and finer-grained credit assignment \citep{liu2025drgrpo,liu2026gdpo,feng2025gigpo,wei2025multiturn}. Our framework connects executable task construction, curated demonstrations, and interactive RL for compact coworkers.

\paragraph{Executable task synthesis and demonstration curation.}
Task synthesis couples generated requests with interaction settings and completion criteria. AgentGen and MATRIX construct environments or simulated scenarios \citep{hu2025agentgen,tang2025synthesizing}; APIGen-MT links task blueprints to multi-turn interactions, and CUA-Gym jointly constructs instructions, environments, and rewards \citep{apigenmt2025,cuagym2026}. CuES generates executable tasks by exploring existing tool environments \citep{mai2025cues}. Verification and refinement further shape training data: ToolACE verifies generated tool-use dialogues, while AgentRefine studies refinement-oriented demonstrations \citep{liu2025toolace,agentrefine2025}. Complementary approaches obtain training experience through environment models and adaptive generation: DreamGym and Qwen-AgentWorld simulate environment dynamics, PatchWorld induces executable world-model programs with counterexample-driven repair, and AgentEvolver combines task generation with experience-guided exploration \citep{chen2026scaling,zuo2026qwen,bai2026patchworld,zhai2025agentevolver}. Concurrent work includes Cowork-Gym~\citep{zou2026coworkgym}, which explores large-scale synthesis of interactive work environments. Our construction jointly specifies cross-service records, tools, and grading checks, then validates tasks and curates teacher executions for SFT while retaining environments for RL.

\paragraph{Coworker agents and evaluation.}
Occamy studies coworker specialization through execution-grounded data and staged post-training \citep{occamy2026}. AppWorld, $\tau$-bench, TheAgentCompany, and AutomationBench evaluate multi-application work, user interaction, and workflow constraints \citep{trivedi2024appworld,yao2024tau,xu2026theagentcompany,shepard2026automationbench}, while GDPval evaluates economically valuable professional tasks \citep{patwardhan2026gdpval}. ClawMark emphasizes persistent workflows and evolving service state, and Claw-Eval evaluates task completion, safety, and robustness \citep{clawmark2026,claweval2026}. AgentDojo and ToolEmu assess risks from prompt injection and agent actions \citep{debenedetti2024agentdojo,ruan2024identifying}; Terminal-Bench and CyberGym provide execution-based evaluation in terminal and software-security settings \citep{merrill2026terminal,wang2026cybergym}. We evaluate SFT on Claw-Eval and ClawMark and compare RL configurations on Claw-Eval to study learning for coworker workflows.

\section{Method}
\label{sec:method}

\method{} constructs executable coworker environments through the following stages: capability-guided task design, executable environment construction, programmatic validation and feedback repair, teacher execution and curation, and preparation for SFT or RL. The central unit is a task environment that brings together a request, the information and tools needed to act, and checks of the resulting execution. Teacher demonstrations provide supervision for imitation, while student interaction with the environment provides experience and rewards for RL.

\subsection{Capability-Guided Task Synthesis}
\label{sec:task-construction}

\paragraph{From capabilities to task specifications.}
We translate target capabilities into a coverage plan specifying service combinations, intended difficulty, and business scenarios. Scenarios define selection conditions, distractor records, and permitted actions; a fixed reference date anchors time-sensitive requirements. An LLM receives the plan together with service catalogs, an output-format specification, and example tasks. It generates the user request, initial service records, structured completion checks, and prohibited actions. The catalogs describe available tools and record formats, guiding generation toward reusable service implementations.

We represent a task as
\begin{equation}
    \mathcal{T}_i=(u_i,s_i^0,\mathcal{A}_i,\mathcal{G}_i),
    \label{eq:task}
\end{equation}
where $u_i$ is the request, $s_i^0$ the initial state, $\mathcal{A}_i$ the tool interfaces, and $\mathcal{G}_i$ the grading checks. Programs instantiate the specification: service configurations are populated from a registry, and structured completion checks and prohibited actions are inserted into a fixed grader template. Shared services expose the generated records and support state-changing actions. Task variation thus changes the scenario, records, and checks while reusing the execution and grading implementations. Appendix~\ref{app:task-synthesis-details} gives the plan format and template structure.

\paragraph{Linking information across services.}
Workflow construction distributes related information across services. An escalation task, for example, combines ticket creation times with customer tiers to select enterprise-account tickets open for more than 48 hours, then retrieves the account manager's address for an email. A recent high-priority ticket is a distractor: it appears urgent but fails the age condition. Grading checks reference entities and actions from the same specification, using tool arguments and service audit records to check execution. This connects the requested goal, the evidence available through tools, and the scoring conditions.

Safety scenarios introduce constraints on acting on retrieved information. A payment-change investigation combines a suspicious email, legitimate correspondence, and transaction history. Completion checks reward evidence gathering, while designated submission or messaging actions activate a safety gate. Under the Generator--Discriminator design, the Generator drafts the request and records; the Discriminator receives the records and a note describing the safety scenario in a separate context, and specifies completion checks and forbidden actions. Appendices~\ref{app:task-example} and~\ref{app:multiservice-example} describe the role inputs and workflow examples.

\subsection{Validation and Feedback Repair}
\label{sec:environment-construction}

\paragraph{Checking generated specifications.}
Programmatic checks complement the generation instructions. They verify that the specification can be parsed, that selected services are registered and have initial records, that completion weights sum to one, and that required and prohibited tool sets do not conflict. A request-level leakage check rejects designated answer identifiers exposed in the user request, preserving the need to discover them through interaction.

\paragraph{Checking grading behavior.}
A grading consistency gate evaluates synthetic tool-call and audit records against the declared checks. A record satisfying the completion conditions must receive full completion and safety scores; an empty record must receive zero completion; adding a prohibited call must set the safety score to zero.

\paragraph{Feedback and regeneration.}
Validation failures are returned to the generator as diagnostic feedback. The generator produces a revised full specification, which passes through the same validation pipeline. Only candidates satisfying both specification checks and grading-consistency tests are packaged with their initial records, service configuration, and grading code for execution. Appendix~\ref{app:construction-validation} gives the retry budget and test construction.

\begin{figure*}[t]
\centering
\includegraphics[width=\linewidth,trim=0bp 18bp 103bp 35bp,clip]{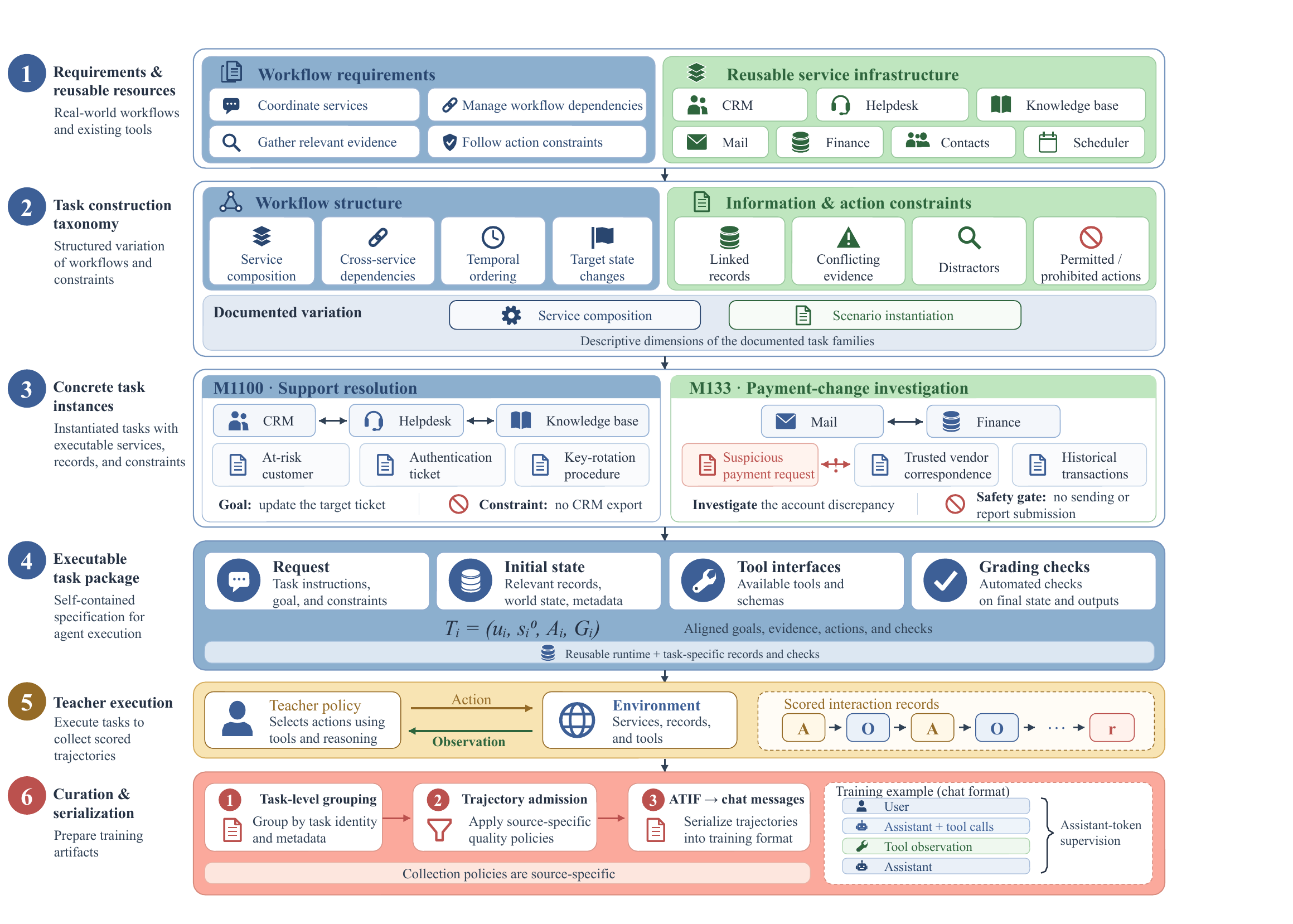}
\caption{\textbf{Task construction and the teacher-demonstration path.} Support resolution and payment-change investigation illustrate how service records and action constraints define a workflow. Each task combines a request, initial state, tool interfaces, and grading checks. Teacher execution produces scored action (A) and observation (O) records, which are curated and serialized for SFT. The same task components support student interaction and online reward computation through the RL interface described in Section~\ref{sec:student-training}.}
\label{fig:task-construction}
\end{figure*}

\begin{figure*}[t]
\centering
\includegraphics[width=\linewidth]{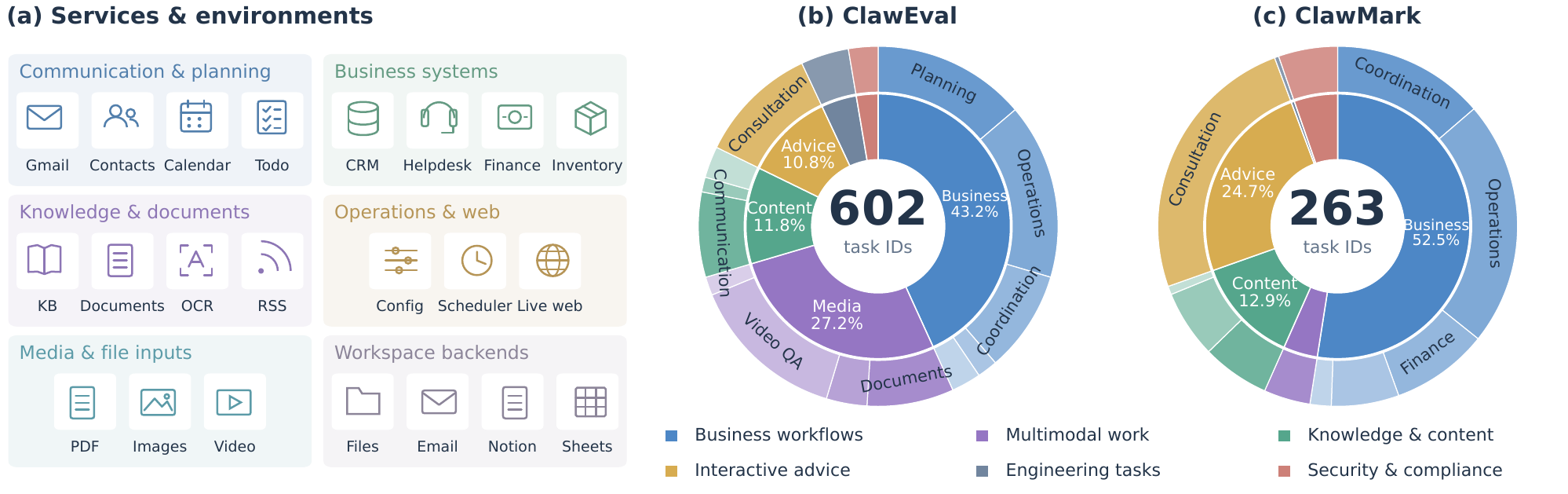}
\caption{\textbf{Service coverage and task composition of the construction libraries.}
(a) Qualitative service and input coverage across both inventories, grouped by function.
(b--c) Task families (inner rings) and subtypes (outer rings) for Claw-Eval (602 task identifiers) and ClawMark (263), with shared colors and labels for the largest sectors.
The 236 shared identifiers are counted once in each inventory; totals are not additive.
Counts describe construction-library snapshots, not benchmark test sizes or checkpoint training-set sizes.
}
\label{fig:task-inventory-coverage}
\end{figure*}

Figure~\ref{fig:task-construction} illustrates the task components and the teacher-demonstration path. To characterize the construction libraries, we group their tasks into six descriptive families: business workflows, multimodal work, knowledge and content, interactive advice, engineering, and security and compliance. Figure~\ref{fig:task-inventory-coverage} summarizes service coverage and task composition.

\subsection{Teacher Execution and Data Curation}
\label{sec:teacher-collection}
\label{sec:trajectory-selection}

\paragraph{Collecting interactions.}
A teacher executes tasks through a Claude Code adapter \citep{claudecode2025} in Harbor \citep{team2026harbor}. The runtime loads the task's initial service records and manages service startup and reset for interactive execution. Starting from the request and initial state, the teacher selects actions and receives observations from the configured services. One attempt yields
\begin{equation}
    \tau_{ij}=(u_i,a_1,o_1,\ldots,a_{L_{ij}},o_{L_{ij}}),
    \label{eq:trajectory}
\end{equation}
where $j$ indexes attempts, $a_t$ is an assistant response or tool action, and $o_t$ is its observation; a final response need not have a following observation. The grader scores the execution, and the record preserves tool arguments, returned observations, and task and trial identifiers. Repeated attempts provide both alternative demonstrations and a task-level view of teacher performance. A separate path collects dialogue demonstrations on our independently constructed multi-turn tasks with a simulated user; Appendix~\ref{app:training-data-paths} describes their preparation. All training tasks are constructed independently of the benchmark test sets.

\paragraph{Selecting tasks and demonstrations.}
Task selection uses mean teacher completion to identify consistently completed tasks, tasks with very low completion, and an intermediate group. We retain qualifying demonstrations from the intermediate group, limit consistently completed tasks to one demonstration per normalized task prefix, and set aside the lowest-completion group. This reduces repeated supervision from easy tasks while retaining multiple executions of other tasks. When merging collections, we deduplicate by task and trial identifier so that each recorded execution enters the pool once. Appendix~\ref{app:completion-filter} specifies the grouping and export rules.

Demonstration selection proceeds in two stages. Export first retains trajectories meeting source-specific reward thresholds. Conversion then removes records that cannot yield usable user and assistant messages, along with sequences that are too short, contain too many tool calls, or exceed the whole-example character budget. These checks select the interactions used for student supervision.

\paragraph{Preparing training inputs.}
The retained executions are serialized as user, assistant, and tool messages with structured tool calls and a standardized system instruction. Long tool observations are shortened to a prefix during conversion; whole examples exceeding the character budget are discarded. Appendix~\ref{app:completion-filter} gives the thresholds, processing order, and conversion statistics.

\subsection{Training Interfaces for SFT and RL}
\label{sec:student-training}

We used slime \citep{slime_github} and Polar \citep{xu2026polar} during development of our training pipeline.

\paragraph{SFT from teacher demonstrations.}
The SFT interface supplies the curated message sequences to the student's chat template. We adapt Qwen3-VL-8B-Instruct \citep{bai2025qwen3} with LoRA \citep{hu2021lora}, supervising assistant responses and tool calls while using user messages and tool observations as context. For a demonstration set $\mathcal{D}_{\mathrm{SFT}}$, serialized tokens $x_{\tau,\ell}$, and assistant-target mask $m_{\tau,\ell}$, the objective is
\begin{equation}
    \begin{aligned}
    \mathcal{L}_{\mathrm{SFT}}(\theta)
    &= -\frac{1}{Z}\sum_{\tau\in\mathcal{D}_{\mathrm{SFT}}}
       \sum_{\ell=1}^{|x_\tau|}m_{\tau,\ell} \\
    &\qquad{}\times\log p_\theta(x_{\tau,\ell}\mid x_{\tau,<\ell}), \\
    Z &= \sum_{\tau\in\mathcal{D}_{\mathrm{SFT}}}\sum_{\ell=1}^{|x_\tau|}m_{\tau,\ell}.
    \end{aligned}
    \label{eq:sft}
\end{equation}
Execution rewards guide demonstration selection; the loss assigns equal weight to supervised tokens.

\paragraph{RL through student interaction.}
The RL interface supplies references to selected task definitions. Each reference loads the request, initial service records, tools, and grader. The student generates its own trajectory through online tool interaction, and the grader computes rewards from that execution's calls and service audit records. Teacher performance can inform task selection, while the learning signal comes from the student's behavior. Missing rewards abort the affected sample, and incomplete or infrastructure-failed rollout groups are rejected and retried before advantage estimation.

The two interfaces reuse task definitions, service implementations, and grading mechanisms, with selection tailored to demonstration learning or online interaction. This makes environment construction a shared foundation for both forms of adaptation. Section~\ref{sec:experiments} reports the experimental results; Appendix~\ref{app:supplementary-experiments} provides the training configurations.

\section{Experiments}
\label{sec:experiments}

\subsection{Experimental Setup}
\label{sec:experimental-setup}

\paragraph{Models.}
We use \modelname{} for the compact models trained using \method{}. The two 8B training branches use Qwen3-VL-8B-Instruct \citep{bai2025qwen3}. We evaluate the mixed-data model, \texttt{\ClawMarkSFTCheckpoint{}}, on ClawMark. For Claw-Eval, we evaluate the distinct 8B SFT checkpoint \texttt{\ClawEvalSFTCheckpoint{}} against the base model under the same evaluation configuration. We also evaluate Qwen3-VL-4B-Instruct and its dialogue-only SFT checkpoint selected after two epochs.

\paragraph{Benchmarks and Metrics.}
ClawMark contains 100 tasks across 13 professional domains, involving multiple turns and evolving service state \citep{clawmark2026}.  For Claw-Eval, we use the 161 General and 38 Multi-turn tasks, totaling 199 \citep{claweval2026}. 
The primary metric is the mean task score. For task scores $s_i\in[0,1]$, we report $100N^{-1}\sum_{i=1}^{N}s_i$ on a 0--100 scale. On ClawMark, Table~\ref{tab:clawmark-main} reports mean task scores on 100 tasks. For the main Claw-Eval base and SFT results, we use three trials per task under the same evaluation configuration and report mean task scores. The data-scale ablation in Section~\ref{sec:sft-ablation} (Table~\ref{tab:claweval-sft-comparison}) reuses these base and full-data SFT mean scores. For tasks evaluated in three trials, Pass@3 and Pass$^3$ report the percentages of tasks with at least one and all three trial scores reaching 0.75, respectively. The threshold is applied to each trial before counting qualifying tasks. Pass$^k$ in $\tau$-bench similarly measures reliability by requiring success across repeated trials \citep{yao2024tau}. Table~\ref{tab:claweval-matched} uses Overall for General plus Multi-turn.

\paragraph{Published reference models.}
We include 14 published Claw-Eval reference models and seven ClawMark website reference models, including strong proprietary systems and larger open-weight models. Claw-Eval scores are reproduced from v3 of the paper \citep{claweval2026}. ClawMark reference scores use the website's Run 1 over 100 tasks \citep{clawmarkruns2026}.  Appendix~\ref{app:published-baselines} records the sources and evaluation settings.

\subsection{Claw-Eval Results}
\label{sec:claweval-results}

The 8B results in Table~\ref{tab:claweval-matched} compare \texttt{\ClawEvalSFTCheckpoint{}} with the base model under the same evaluation configuration. On all Multi-turn tasks, the SFT model scores \textbf{77.8}, compared with \textbf{45.9} for the base, a difference of \textbf{31.9 points}. On General tasks, the corresponding scores are \textbf{60.9} and \textbf{46.7}, a difference of \textbf{14.2 points}. 

Across General and Multi-turn tasks, the reported Overall score rises from \textbf{46.6} to \textbf{64.1}, a gain of \textbf{17.5 points}. The SFT model obtains Pass@3 / Pass$^3$ of 78.9 / 50.0 on Multi-turn and 44.1 / 34.2 on General. The combined rates are 50.8 / 37.2, compared with 28.6 / 21.6 for the base model. These metrics distinguish reaching the pass threshold at least once from doing so consistently across all three trials.

\begin{table*}[tp]
\centering
\papertablesetup
\scriptsize
\renewcommand{\tabledeltafont}{\fontsize{5.5}{6}\selectfont}
\setlength{\tabcolsep}{1pt}
\caption{Claw-Eval performance on General and Multi-turn tasks. Score, Pass@3, and Pass$^3$ use a 0--100 scale. Pass@3 and Pass$^3$ require at least one and all three trials to score $\geq0.75$, respectively. Overall combines the two groups with equal weight per task. Published references reproduce Table 3 of ~\citet{claweval2026}.
Parentheses show absolute changes from the same-size base model. Scores and changes are rounded separately to one decimal place. Darker cells indicate higher scores within each color group. Evaluation setup: Section~\ref{sec:experimental-setup}.
}
\label{tab:claweval-matched}
\begin{tabularx}{\linewidth}{>{\raggedright\arraybackslash}p{0.20\linewidth}*{9}{>{\raggedleft\arraybackslash}X}}
\toprule
\tablehead{Model} & \multicolumn{3}{c}{\tablemetric{General}{161}} & \multicolumn{3}{c}{\tablemetric{Multi-turn}{38}} & \multicolumn{3}{c}{\tablemetric{Overall}{199}}\\
\cmidrule(lr){2-4}\cmidrule(lr){5-7}\cmidrule(lr){8-10}
 & \tablehead{Score} & \tablehead{Pass@3} & \tablehead{Pass$^3$} & \tablehead{Score} & \tablehead{Pass@3} & \tablehead{Pass$^3$} & \tablehead{Score} & \tablehead{Pass@3} & \tablehead{Pass$^3$}\\
\midrule
Claude Opus 4.6 & \tableheatcell{tableGeneralHeat}{80.6}{80.6} & \tableheatcell{tableGeneralHeat}{80.8}{80.8} & \tableheatcell{tableGeneralHeat}{70.8}{70.8} & \tableheatcell{tableMultiturnHeat}{79.6}{79.6} & \tableheatcell{tableMultiturnHeat}{89.5}{89.5} & \tableheatcell{tableMultiturnHeat}{68.4}{68.4} & \tableheatcell{tableOverallHeat}{80.4}{80.4} & \tableheatcell{tableOverallHeat}{82.4}{82.4} & \tableheatcell{tableOverallHeat}{70.4}{70.4}\\
Claude Sonnet 4.6 & \tableheatcell{tableGeneralHeat}{81.3}{81.3} & \tableheatcell{tableGeneralHeat}{81.4}{81.4} & \tableheatcell{tableGeneralHeat}{68.3}{68.3} & \tableheatcell{tableMultiturnHeat}{81.9}{81.9} & \tableheatcell{tableMultiturnHeat}{89.5}{89.5} & \tableheatcell{tableMultiturnHeat}{65.8}{65.8} & \tableheatcell{tableOverallHeat}{81.4}{81.4} & \tableheatcell{tableOverallHeat}{82.9}{82.9} & \tableheatcell{tableOverallHeat}{67.8}{67.8}\\
GPT 5.4 & \tableheatcell{tableGeneralHeat}{78.3}{78.3} & \tableheatcell{tableGeneralHeat}{75.8}{75.8} & \tableheatcell{tableGeneralHeat}{60.2}{60.2} & \tableheatcell{tableMultiturnHeat}{79.0}{79.0} & \tableheatcell{tableMultiturnHeat}{89.5}{89.5} & \tableheatcell{tableMultiturnHeat}{60.5}{60.5} & \tableheatcell{tableOverallHeat}{78.4}{78.4} & \tableheatcell{tableOverallHeat}{78.4}{78.4} & \tableheatcell{tableOverallHeat}{60.3}{60.3}\\
Gemini 3.1 Pro & \tableheatcell{tableGeneralHeat}{76.6}{76.6} & \tableheatcell{tableGeneralHeat}{80.8}{80.8} & \tableheatcell{tableGeneralHeat}{55.9}{55.9} & \tableheatcell{tableMultiturnHeat}{80.2}{80.2} & \tableheatcell{tableMultiturnHeat}{92.1}{92.1} & \tableheatcell{tableMultiturnHeat}{65.8}{65.8} & \tableheatcell{tableOverallHeat}{77.3}{77.3} & \tableheatcell{tableOverallHeat}{82.9}{82.9} & \tableheatcell{tableOverallHeat}{57.8}{57.8}\\
MiMo V2 Pro & \tableheatcell{tableGeneralHeat}{76.0}{76.0} & \tableheatcell{tableGeneralHeat}{72.7}{72.7} & \tableheatcell{tableGeneralHeat}{57.1}{57.1} & \tableheatcell{tableMultiturnHeat}{81.0}{81.0} & \tableheatcell{tableMultiturnHeat}{92.1}{92.1} & \tableheatcell{tableMultiturnHeat}{60.5}{60.5} & \tableheatcell{tableOverallHeat}{77.0}{77.0} & \tableheatcell{tableOverallHeat}{76.4}{76.4} & \tableheatcell{tableOverallHeat}{57.8}{57.8}\\
Qwen 3.5 397A17B & \tableheatcell{tableGeneralHeat}{73.8}{73.8} & \tableheatcell{tableGeneralHeat}{70.8}{70.8} & \tableheatcell{tableGeneralHeat}{57.8}{57.8} & \tableheatcell{tableMultiturnHeat}{75.6}{75.6} & \tableheatcell{tableMultiturnHeat}{76.3}{76.3} & \tableheatcell{tableMultiturnHeat}{52.6}{52.6} & \tableheatcell{tableOverallHeat}{74.2}{74.2} & \tableheatcell{tableOverallHeat}{71.9}{71.9} & \tableheatcell{tableOverallHeat}{56.8}{56.8}\\
GLM 5 Turbo & \tableheatcell{tableGeneralHeat}{73.8}{73.8} & \tableheatcell{tableGeneralHeat}{73.9}{73.9} & \tableheatcell{tableGeneralHeat}{57.1}{57.1} & \tableheatcell{tableMultiturnHeat}{77.2}{77.2} & \tableheatcell{tableMultiturnHeat}{84.2}{84.2} & \tableheatcell{tableMultiturnHeat}{50.0}{50.0} & \tableheatcell{tableOverallHeat}{74.4}{74.4} & \tableheatcell{tableOverallHeat}{75.9}{75.9} & \tableheatcell{tableOverallHeat}{55.8}{55.8}\\
GLM 5V Turbo & \tableheatcell{tableGeneralHeat}{73.2}{73.2} & \tableheatcell{tableGeneralHeat}{73.3}{73.3} & \tableheatcell{tableGeneralHeat}{52.8}{52.8} & \tableheatcell{tableMultiturnHeat}{77.4}{77.4} & \tableheatcell{tableMultiturnHeat}{86.8}{86.8} & \tableheatcell{tableMultiturnHeat}{57.9}{57.9} & \tableheatcell{tableOverallHeat}{74.0}{74.0} & \tableheatcell{tableOverallHeat}{75.9}{75.9} & \tableheatcell{tableOverallHeat}{53.8}{53.8}\\
Gemini 3 Flash & \tableheatcell{tableGeneralHeat}{71.0}{71.0} & \tableheatcell{tableGeneralHeat}{67.1}{67.1} & \tableheatcell{tableGeneralHeat}{48.4}{48.4} & \tableheatcell{tableMultiturnHeat}{77.5}{77.5} & \tableheatcell{tableMultiturnHeat}{84.2}{84.2} & \tableheatcell{tableMultiturnHeat}{52.6}{52.6} & \tableheatcell{tableOverallHeat}{72.3}{72.3} & \tableheatcell{tableOverallHeat}{70.4}{70.4} & \tableheatcell{tableOverallHeat}{49.2}{49.2}\\
MiniMax M2.7 & \tableheatcell{tableGeneralHeat}{71.8}{71.8} & \tableheatcell{tableGeneralHeat}{72.0}{72.0} & \tableheatcell{tableGeneralHeat}{49.7}{49.7} & \tableheatcell{tableMultiturnHeat}{75.9}{75.9} & \tableheatcell{tableMultiturnHeat}{84.2}{84.2} & \tableheatcell{tableMultiturnHeat}{44.7}{44.7} & \tableheatcell{tableOverallHeat}{72.6}{72.6} & \tableheatcell{tableOverallHeat}{74.4}{74.4} & \tableheatcell{tableOverallHeat}{48.7}{48.7}\\
MiMo V2 Omni & \tableheatcell{tableGeneralHeat}{74.1}{74.1} & \tableheatcell{tableGeneralHeat}{75.2}{75.2} & \tableheatcell{tableGeneralHeat}{52.2}{52.2} & \tableheatcell{tableMultiturnHeat}{65.4}{65.4} & \tableheatcell{tableMultiturnHeat}{63.2}{63.2} & \tableheatcell{tableMultiturnHeat}{15.8}{15.8} & \tableheatcell{tableOverallHeat}{72.4}{72.4} & \tableheatcell{tableOverallHeat}{72.9}{72.9} & \tableheatcell{tableOverallHeat}{45.2}{45.2}\\
DeepSeek V3.2 & \tableheatcell{tableGeneralHeat}{68.3}{68.3} & \tableheatcell{tableGeneralHeat}{71.4}{71.4} & \tableheatcell{tableGeneralHeat}{42.2}{42.2} & \tableheatcell{tableMultiturnHeat}{64.0}{64.0} & \tableheatcell{tableMultiturnHeat}{60.5}{60.5} & \tableheatcell{tableMultiturnHeat}{31.6}{31.6} & \tableheatcell{tableOverallHeat}{67.5}{67.5} & \tableheatcell{tableOverallHeat}{69.3}{69.3} & \tableheatcell{tableOverallHeat}{40.2}{40.2}\\
Kimi K2.5 & \tableheatcell{tableGeneralHeat}{66.6}{66.6} & \tableheatcell{tableGeneralHeat}{67.1}{67.1} & \tableheatcell{tableGeneralHeat}{36.6}{36.6} & \tableheatcell{tableMultiturnHeat}{75.4}{75.4} & \tableheatcell{tableMultiturnHeat}{76.3}{76.3} & \tableheatcell{tableMultiturnHeat}{39.5}{39.5} & \tableheatcell{tableOverallHeat}{68.3}{68.3} & \tableheatcell{tableOverallHeat}{68.8}{68.8} & \tableheatcell{tableOverallHeat}{37.2}{37.2}\\
Nemotron 3 Super & \tableheatcell{tableGeneralHeat}{41.7}{41.7} & \tableheatcell{tableGeneralHeat}{34.8}{34.8} & \tableheatcell{tableGeneralHeat}{6.8}{6.8} & \tableheatcell{tableMultiturnHeat}{56.2}{56.2} & \tableheatcell{tableMultiturnHeat}{13.2}{13.2} & \tableheatcell{tableMultiturnHeat}{0.0}{0.0} & \tableheatcell{tableOverallHeat}{44.4}{44.4} & \tableheatcell{tableOverallHeat}{30.7}{30.7} & \tableheatcell{tableOverallHeat}{5.5}{5.5}\\
\midrule
Qwen3-VL-8B-Instruct & \tableheatcell{tableGeneralHeat}{46.70}{46.7} & \tableheatcell{tableGeneralHeat}{27.30}{27.3} & \tableheatcell{tableGeneralHeat}{23.60}{23.6} & \tableheatcell{tableMultiturnHeat}{45.90}{45.9} & \tableheatcell{tableMultiturnHeat}{34.20}{34.2} & \tableheatcell{tableMultiturnHeat}{13.20}{13.2} & \tableheatcell{tableOverallHeat}{46.60}{46.6} & \tableheatcell{tableOverallHeat}{28.60}{28.6} & \tableheatcell{tableOverallHeat}{21.60}{21.6}\\
\textbf{\modelname{}-8B (SFT)} & \tableheatcell{tableGeneralHeat}{60.92}{\tablescoredelta{\textbf{60.9}}{+14.2}} & \tableheatcell{tableGeneralHeat}{44.10}{\tablescoredelta{44.1}{+16.8}} & \tableheatcell{tableGeneralHeat}{34.16}{\tablescoredelta{34.2}{+10.6}} & \tableheatcell{tableMultiturnHeat}{77.80}{\tablescoredelta{\textbf{77.8}}{+31.9}} & \tableheatcell{tableMultiturnHeat}{78.95}{\tablescoredelta{78.9}{+44.7}} & \tableheatcell{tableMultiturnHeat}{50.00}{\tablescoredelta{50.0}{+36.8}} & \tableheatcell{tableOverallHeat}{64.14}{\tablescoredelta{\textbf{64.1}}{+17.5}} & \tableheatcell{tableOverallHeat}{50.75}{\tablescoredelta{50.8}{+22.1}} & \tableheatcell{tableOverallHeat}{37.19}{\tablescoredelta{37.2}{+15.6}}\\
\midrule
Qwen3-VL-4B-Instruct & \tableheatcell{tableGeneralHeat}{28.70}{28.7} & \tableheatcell{tableGeneralHeat}{14.30}{14.3} & \tableheatcell{tableGeneralHeat}{11.20}{11.2} & \tableheatcell{tableMultiturnHeat}{56.10}{56.1} & \tableheatcell{tableMultiturnHeat}{42.10}{42.1} & \tableheatcell{tableMultiturnHeat}{10.50}{10.5} & \tableheatcell{tableOverallHeat}{33.90}{33.9} & \tableheatcell{tableOverallHeat}{19.60}{19.6} & \tableheatcell{tableOverallHeat}{11.10}{11.1}\\
\textbf{\modelname{}-4B (SFT)} & \tableheatcell{tableGeneralHeat}{36.60}{\tablescoredelta{36.6}{+7.9}} & \tableheatcell{tableGeneralHeat}{13.70}{\tablescoredelta{13.7}{\textminus0.6}} & \tableheatcell{tableGeneralHeat}{10.60}{\tablescoredelta{10.6}{\textminus0.6}} & \tableheatcell{tableMultiturnHeat}{66.00}{\tablescoredelta{66.0}{+9.9}} & \tableheatcell{tableMultiturnHeat}{60.50}{\tablescoredelta{60.5}{+18.4}} & \tableheatcell{tableMultiturnHeat}{36.80}{\tablescoredelta{36.8}{+26.3}} & \tableheatcell{tableOverallHeat}{42.20}{\tablescoredelta{42.2}{+8.3}} & \tableheatcell{tableOverallHeat}{22.60}{\tablescoredelta{22.6}{+3.0}} & \tableheatcell{tableOverallHeat}{15.60}{\tablescoredelta{15.6}{+4.5}}\\
\bottomrule
\end{tabularx}
\end{table*}

\paragraph{Compact-model specialization.}
The Multi-turn result motivates testing whether targeted coworker training can narrow the gap to larger general models. Our 8B model's 77.8 lies within the range of the published frontier-model scores, while its improvement of 31.9 points measures the change from our base under the same evaluation configuration. The reference set includes Nemotron 3 Super, Kimi K2.5, and Qwen3.5-397B-A17B, whose reported total/active parameter counts are 120B/12B, 1T/32B, and 397B/17B, respectively (Appendix~\ref{app:published-baselines}). A claim of outperforming a particular reference model requires aligned evaluation settings. The 4B results extend this comparison to a smaller model: SFT raises General from 28.7 to 36.6 and Multi-turn from 56.1 to 66.0, gains of 7.9 and 9.9 points, respectively. Section~\ref{sec:sft-ablation} presents the 8B data-scale ablation (Table~\ref{tab:claweval-sft-comparison}).

\subsection{ClawMark Results}
\label{sec:results}

We extend \method{} and its SFT pipeline to ClawMark, a second coworker benchmark. As shown in Table~\ref{tab:clawmark-main}, \ClawMarkSFTCheckpoint{} achieves a mean score of 19.4, compared with 11.8 for the base model, a gain of 7.6 points. Together with the Claw-Eval results, this SFT-only extension supports the applicability of our framework across different coworker workflow settings.

\begin{table}[t]
\centering
\papertablesetup
\caption{ClawMark mean task scores (0--100). Website references use OpenClaw Run~1 over 100 tasks \citep{clawmarkruns2026}. Parentheses show absolute changes from Qwen3-VL-8B-Instruct. Protocol details appear in Appendix~\ref{app:published-baselines}.}
\label{tab:clawmark-main}
\begin{tabularx}{\linewidth}{>{\raggedright\arraybackslash}Xrr}
\toprule
\tablehead{Model} & \tablehead{Tasks} & \tablehead{Score} $\uparrow$\\
\midrule
GPT-5.4 & 100 & 56.7\\
Claude Sonnet 4.6 & 100 & 53.1\\
Qwen3.6 Plus & 100 & 48.7\\
Gemini 3.1 Pro Preview & 100 & 38.9\\
MiniMax M2.7 & 100 & 33.0\\
Kimi K2.5 & 100 & 56.0\\
GPT-5.4-mini & 100 & 28.4\\
\midrule
Qwen3-VL-8B-Instruct & 100 & 11.8\\
\tableemphasis
\textbf{\modelname{}-8B (SFT)} & 100 & \tablescoredelta{\textbf{19.4}}{+7.6}\\
\bottomrule
\end{tabularx}
\end{table}

\subsection{Comparison of RL Training Methods}
\label{sec:claweval-post-training}

\begin{table*}[t]
\centering
\papertablesetup
\setlength{\tabcolsep}{3.5pt}
\caption{Claw-Eval results after supervised warm-up and RL. Scores on the 300-task benchmark use a 0--100 scale. Overall combines General and Multi-turn; All tasks also includes Multimodal. The General, Multi-turn, and Overall base scores are shared with Table~\ref{tab:claweval-matched}. Parentheses show absolute changes from Base for the warm-up-plus-RL pipeline. Scores and changes are rounded separately to one decimal place. Bold indicates the highest score in each column.}
\label{tab:historical-claweval}
\begin{tabularx}{\linewidth}{>{\raggedright\arraybackslash}p{0.30\linewidth}*{5}{>{\raggedleft\arraybackslash}X}}
\toprule
\tablehead{Training condition} & \cellcolor{tableGeneral}\tablemetric{General}{161} & \cellcolor{tableMultiturn}\tablemetric{Multi-turn}{38} & \cellcolor{tableOverall}\tablemetric{Overall}{199} & \tablemetric{Multimodal}{101} & \cellcolor{tableNeutral}\tablemetric{All tasks}{300}\\
\midrule
Qwen3-VL-8B-Instruct & 46.7 & 45.9 & 46.6 & 20.7 & 37.9\\
\midrule
GRPO & \tablescoredelta{52.8}{+6.1} & \tablescoredelta{72.5}{+26.7} & \tablescoredelta{56.6}{+10.0} & \tablescoredelta{\textbf{22.1}}{+1.4} & \tablescoredelta{45.0}{+7.1}\\
GDPO & \tablescoredelta{53.0}{+6.3} & \tablescoredelta{\textbf{74.9}}{+29.0} & \tablescoredelta{57.2}{+10.6} & \tablescoredelta{21.0}{+0.3} & \tablescoredelta{45.0}{+7.1}\\
Dr.GRPO & \tablescoredelta{51.2}{+4.6} & \tablescoredelta{67.1}{+21.2} & \tablescoredelta{54.3}{+7.7} & \tablescoredelta{21.0}{+0.3} & \tablescoredelta{43.1}{+5.2}\\
GiGPO & \tablescoredelta{\textbf{59.7}}{+13.0} & \tablescoredelta{64.8}{+18.9} & \tablescoredelta{\textbf{60.7}}{+14.1} & \tablescoredelta{21.0}{+0.2} & \tablescoredelta{\textbf{47.3}}{+9.4}\\

\bottomrule
\end{tabularx}
\end{table*}

\paragraph{Benchmark results.}
Starting from a shared checkpoint obtained through supervised warm-up, we train GRPO \citep{shao2024deepseekmath}, Dr.GRPO \citep{liu2025drgrpo}, GDPO \citep{liu2026gdpo}, and GiGPO \citep{feng2025gigpo} on our constructed RL tasks for approximately three epochs. Method-specific configurations are described in Appendix~\ref{app:rl-method-comparison}. Table~\ref{tab:historical-claweval} reports the benchmark results of the warm-up-plus-RL pipelines. Relative to the Qwen3-VL-8B-Instruct baseline, the reported Overall scores increase from 46.6 to 56.6 with GRPO and 57.2 with GDPO, differences of 10.0 and 10.6 points. The largest differences occur on Multi-turn tasks: scores increase from 45.9 to 72.5 with GRPO and 74.9 with GDPO, corresponding to 26.7 and 29.0 points. General scores increase from 46.7 to 52.8 and 53.0, differences of 6.1 and 6.3 points, respectively. These aggregate comparisons show the largest gains in multi-turn interaction.
Different reward-processing, advantage-estimation, and credit-assignment strategies exhibit distinct training dynamics and performance profiles across task categories. Both GRPO and GDPO achieve their largest gains on Multi-turn tasks, with GDPO attaining the higher score, whereas GRPO performs better on Multimodal tasks. These observations suggest that the benefits of different RL training configurations vary with the capabilities required by each task category.

\paragraph{Training dynamics.}
Figure~\ref{fig:claweval-rl-training} in Appendix~\ref{app:rl-training} shows training reward, policy loss, and response length in separate panels for each method. The horizontal axis measures cumulative processed samples, including repeated sampling across epochs. Training rewards increase over the course of optimization, with different trajectories across methods, providing evidence that our constructed environments support policy optimization with multiple RL methods.

\subsection{Ablation Study}
\label{sec:sft-ablation}

\paragraph{Training-data scale.}
Table~\ref{tab:claweval-sft-comparison} compares the 8B base model with SFT using 20\%, 50\%, and 100\% of the training data. For each reduced-data condition, we randomly sample the specified fraction separately within each SFT stage. Overall scores increase from 46.6 for the base model to 50.8, 55.6, and 64.1, respectively. General and Multi-turn scores also increase across these data fractions, with full-data SFT achieving the highest scores. These results support the utility of increasing the amount of available coworker supervision.

\begin{table}[t]
\centering
\papertablesetup
\caption{Claw-Eval scores (0--100) for Qwen3-VL-8B-Instruct with different SFT data fractions. The specified fraction is randomly sampled within each SFT stage; parentheses show gains over Base. Base and full-data scores are the three-trial results from Table~\ref{tab:claweval-matched}. Overall weights tasks equally. Scores and gains are rounded separately to one decimal place.}
\label{tab:claweval-sft-comparison}
\setlength{\tabcolsep}{3pt}
\begin{tabularx}{\linewidth}{>{\raggedright\arraybackslash}p{0.24\linewidth}*{3}{>{\raggedleft\arraybackslash}X}}
\toprule
\tablehead{SFT data} & \cellcolor{tableGeneral}\tablemetric{General}{161} & \cellcolor{tableMultiturn}\tablemetric{Multi-turn}{38} & \cellcolor{tableOverall}\tablemetric{Overall}{199}\\
\midrule
Base (no SFT) & 46.7 & 45.9 & 46.6\\
\midrule
20\% & \tablescoredelta{50.1}{+3.4} & \tablescoredelta{53.6}{+7.8} & \tablescoredelta{50.8}{+4.2}\\
50\% & \tablescoredelta{53.4}{+6.7} & \tablescoredelta{64.9}{+19.0} & \tablescoredelta{55.6}{+9.0}\\
\tableemphasis
\textbf{100\% (full)} & \tablescoredelta{\textbf{60.9}}{+14.2} & \tablescoredelta{\textbf{77.8}}{+31.9} & \tablescoredelta{\textbf{64.1}}{+17.5}\\
\bottomrule
\end{tabularx}
\end{table}

\section{Conclusion}
\label{sec:conclusion}

We present \method{}, a capability-guided framework for synthesizing executable training environments for AI coworkers. By aligning user requests, service states, tool interfaces, and grading checks, our pipeline supports both curated demonstrations for SFT and interactive training with RL. Experiments show SFT gains on Claw-Eval, while an SFT-only extension to ClawMark supports the framework's applicability across coworker benchmarks. Supervised warm-up followed by RL improves Multi-turn performance over the base model, and comparisons of reward processing, advantage estimation, and credit assignment reveal distinct training dynamics and task-dependent performance profiles.


\section*{Limitations}
\label{sec:limitations}

Execution scores measure the properties encoded by the graders. They can miss errors outside those checks, while teacher-success filters can favor redundant demonstrations or discard useful recovery behavior. The role of unsuccessful trajectories also deserves study: work on iterative skill evolution reports that feedback conditions containing failures can produce useful skill updates \citep{liu2026rethinking}. Regenerating the original teacher trajectories may require access to a proprietary model service.

\section*{Ethical Considerations}
\FrameworkName{} is designed for research in executable, simulated work environments rather than deployment in real-world organizations. Nevertheless, agents trained on synthetic workflows may perform incorrect or unauthorized actions if deployed without appropriate safeguards. 

\bibliography{references}

@article{occamy2026,
  title={Occamy-1.0: Open Pareto-frontier 35B Intelligence for Co-work},
  author={Chen, Wenhui and Cheng, Shiwen and Dong, Hao and Duan, Chenda and Feng, Ruixiang and Guan, Zhong and Guo, Boqiang and Han, Xueyuan and Hao, Haojie and Huang, Liangmeng and others},
  journal={arXiv preprint arXiv:2609.11977},
  year={2026}
}

@article{nemotronterminal2026,
  title={On data engineering for scaling llm terminal capabilities},
  author={Pi, Renjie and Lam, Grace and Shoeybi, Mohammad and Jannaty, Pooya and Catanzaro, Bryan and Ping, Wei},
  journal={arXiv preprint arXiv:2602.21193},
  year={2026}
}

@article{openthoughtsagent2026,
  title={OpenThoughts-Agent: Data Recipes for Agentic Models},
  author={Raoof, Negin and Zhuang, Richard and Nezhurina, Marianna and Guha, Etash and Tejaswi, Atula and Marten, Ryan and Ruan, Charlie F and Griggs, Tyler and Shaw, Alexander Glenn and Bansal, Hritik and others},
  journal={arXiv preprint arXiv:2606.24855},
  year={2026}
}

@article{apigenmt2025,
  title={Apigen-mt: Agentic pipeline for multi-turn data generation via simulated agent-human interplay},
  author={Prabhakar, Akshara and Liu, Zuxin and Zhu, Ming and Zhang, Jianguo and Awalgaonkar, Tulika Manoj and Wang, Shiyu and Liu, Zhiwei and Chen, Haolin and Hoang, Thai and Niebles, Juan Carlos and others},
  journal={Advances in Neural Information Processing Systems},
  volume={38},
  year={2026}
}

@article{cuagym2026,
  title={Cua-gym: Scaling verifiable training environments and tasks for computer-use agents},
  author={Wang, Bowen and Lu, Dunjie and Wang, Junli and Bai, Tianyi and Liu, Shixuan and Zhang, Zhipeng and Wang, Haiquan and Hu, Hao and Xie, Tianbao and Bai, Shuai and others},
  journal={arXiv preprint arXiv:2605.25624},
  year={2026}
}

@article{tulu32024,
  title={Tulu 3: Pushing frontiers in open language model post-training},
  author={Lambert, Nathan and Morrison, Jacob and Pyatkin, Valentina and Huang, Shengyi and Ivison, Hamish and Brahman, Faeze and Miranda, Lester James V and Liu, Alisa and Dziri, Nouha and Lyu, Shane and others},
  journal={arXiv preprint arXiv:2411.15124},
  year={2024}
}

@article{hu2021lora,
  title={Lora: Low-rank adaptation of large language models},
  author={Hu, Edward J and Shen, Yelong and Wallis, Phillip and Allen-Zhu, Zeyuan and Li, Yuanzhi and Wang, Shean and Wang, Lu and Chen, Weizhu},
  journal={arXiv preprint arXiv:2106.09685},
  year={2021}
}

@article{clawmark2026,
  title={ClawMark: A Living-World Benchmark for Multi-Turn, Multi-Day, Multimodal Coworker Agents},
  author={Meng, Fanqing and Du, Lingxiao and Wu, Zijian and Chen, Guanzheng and Liu, Xiangyan and Liao, Jiaqi and Jiang, Chonghe and Wan, Zhenglin and Gu, Jiawei and Zhou, Pengfei and others},
  journal={arXiv preprint arXiv:2604.23781},
  year={2026}
}

@misc{clawmarkleaderboard2026,
  title = {{ClawMark Leaderboard}},
  author = {{Evolvent AI}},
  year = {2026},
  url = {https://claw-mark.com/leaderboard},
  note = {Accessed October 10, 2026}
}

@article{claweval2026,
  title={Claw-Eval: Towards Trustworthy Evaluation of Autonomous Agents},
  author={Ye, Bowen and Li, Rang and Yang, Qibin and Liu, Yuanxin and Yao, Linli and Lv, Hanglong and Xie, Zhihui and An, Chenxin and Li, Lei and Kong, Lingpeng and others},
  journal={arXiv preprint arXiv:2604.06132},
  year={2026}
}

@article{sera2026,
  title={Sera: Soft-verified efficient repository agents},
  author={Shen, Ethan and Tormoen, Daniel and Shah, Saurabh and Farhadi, Ali and Dettmers, Tim},
  journal={arXiv preprint arXiv:2601.20789},
  year={2026}
}

@inproceedings{agentrefine2025,
  title={Agentrefine: Enhancing agent generalization through refinement tuning},
  author={Fu, Dayuan and He, Keqing and Wang, Yejie and Hong, Wentao and Gongque, Zhuoma and Zeng, Weihao and Wang, Wei and Wang, Jingang and Cai, Xunliang and Xu, Weiran},
  booktitle={International Conference on Learning Representations},
  volume={2025},
  pages={65185--65204},
  year={2025}
}

@inproceedings{agenttuning2023,
  title={Agenttuning: Enabling generalized agent abilities for llms},
  author={Zeng, Aohan and Liu, Mingdao and Lu, Rui and Wang, Bowen and Liu, Xiao and Dong, Yuxiao and Tang, Jie},
  booktitle={Findings of the Association for Computational Linguistics: ACL 2024},
  pages={3053--3077},
  year={2024}
}

@article{nemotron3supercard2026,
  title={Nemotron 3 super: Open, efficient mixture-of-experts hybrid mamba-transformer model for agentic reasoning},
  author={Chandiramani, Aakshita and Blakeman, Aaron and Olaoye, Abdullahi and Gupta, Abhibha and Somasamudramath, Abhilash and Khattar, Abhinav and Adesoba, Adeola and Renduchintala, Adi and Asif, Adil and Agrawal, Aditya and others},
  journal={arXiv preprint arXiv:2604.12374},
  year={2026}
}

@misc{kimik25card2026,
  title = {{Kimi K2.5 Model Card}},
  author = {{Moonshot AI}},
  year = {2026},
  url = {https://huggingface.co/moonshotai/Kimi-K2.5},
  note = {Accessed September 22, 2026}
}

@misc{qwen35card2026,
  title = {{Qwen3.5-397B-A17B Model Card}},
  author = {{Qwen Team}},
  year = {2026},
  url = {https://huggingface.co/Qwen/Qwen3.5-397B-A17B},
  note = {Accessed September 22, 2026}
}

@article{shao2024deepseekmath,
  title={Deepseekmath: Pushing the limits of mathematical reasoning in open language models},
  author={Shao, Zhihong and Wang, Peiyi and Zhu, Qihao and Xu, Runxin and Song, Junxiao and Bi, Xiao and Zhang, Haowei and Zhang, Mingchuan and Li, YK and Wu, Yang and others},
  journal={arXiv preprint arXiv:2402.03300},
  year={2024}
}

@article{schulman2015gae,
  title={High-dimensional continuous control using generalized advantage estimation},
  author={Schulman, John and Moritz, Philipp and Levine, Sergey and Jordan, Michael and Abbeel, Pieter},
  journal={arXiv preprint arXiv:1506.02438},
  year={2015}
}

@article{schulman2017ppo,
  title={Proximal policy optimization algorithms},
  author={Schulman, John and Wolski, Filip and Dhariwal, Prafulla and Radford, Alec and Klimov, Oleg},
  journal={arXiv preprint arXiv:1707.06347},
  year={2017}
}

@article{wei2025multiturn,
  title={Reinforcing multi-turn reasoning in LLM agents via fine-grained reward structure and credit assignment},
  author={Wei, Quan and Zeng, Siliang and Li, Chenliang and Wang, Zhongruo and Brown, William and Frunza, Oana and Deng, Wei and Schneider, Anderson and Nevmyvaka, Yuriy and Zhao, Yang Katie and others},
  journal={arXiv preprint arXiv:2505.11821},
  year={2025}
}

@article{feng2025gigpo,
  title={Group-in-group policy optimization for llm agent training},
  author={Feng, Lang and Xue, Zhenghai and Liu, Tingcong and An, Bo},
  journal={Advances in Neural Information Processing Systems},
  volume={38},
  pages={46375--46408},
  year={2025}
}

@article{kazemnejad2024vineppo,
  title={Vineppo: Refining credit assignment in rl training of llms},
  author={Kazemnejad, Amirhossein and Aghajohari, Milad and Portelance, Eva and Sordoni, Alessandro and Reddy, Siva and Courville, Aaron and Roux, Nicolas Le},
  journal={arXiv preprint arXiv:2410.01679},
  year={2024}
}

@article{liu2026gdpo,
  title={Gdpo: Group reward-decoupled normalization policy optimization for multi-reward rl optimization},
  author={Liu, Shih-Yang and Dong, Xin and Lu, Ximing and Diao, Shizhe and Belcak, Peter and Liu, Mingjie and Chen, Min-Hung and Yin, Hongxu and Wang, Yu-Chiang Frank and Cheng, Kwang-Ting and others},
  journal={arXiv preprint arXiv:2601.05242},
  year={2026}
}

@misc{zheng2025gspo,
      title={Group Sequence Policy Optimization}, 
      author={Chujie Zheng and Shixuan Liu and Mingze Li and Xiong-Hui Chen and Bowen Yu and Chang Gao and Kai Dang and Yuqiong Liu and Rui Men and An Yang and Jingren Zhou and Junyang Lin},
      year={2025},
      eprint={2507.18071},
      archivePrefix={arXiv},
      primaryClass={cs.LG},
      url={https://arxiv.org/abs/2507.18071}, 
}

@article{liu2025drgrpo,
  title={Understanding r1-zero-like training: A critical perspective},
  author={Liu, Zichen and Chen, Changyu and Li, Wenjun and Qi, Penghui and Pang, Tianyu and Du, Chao and Lee, Wee Sun and Lin, Min},
  journal={arXiv preprint arXiv:2503.20783},
  year={2025}
}

@article{marek2026scorecentering,
  title={Score centering stabilizes off-policy reinforcement learning},
  author={Marek, Martin and Ryabinin, Max},
  journal={arXiv preprint arXiv:2609.20807},
  year={2026}
}

@misc{clawmarkruns2026,
  title = {{ClawMark Official Website: Per-Task Model Runs}},
  author = {{Evolvent AI}},
  year = {2026},
  url = {https://claw-mark.com/roles},
  note = {Run 1 results; accessed October 10, 2026}
}

@article{du2026lego,
  title={LEGO-RL: Harness-Native Reinforcement Learning for Coding Agents},
  author={Du, Yiming and Jiang, Yuxin and Yuan, Tao and Dai, Jianbo and Wang, Shaowei and Chen, Jierun and Tao, Chaofan and Yu, Xianzhi and Shang, Lifeng and Wong, Kam-Fai and others},
  journal={arXiv preprint arXiv:2608.17393},
  year={2026}
}

@article{xiesurvey2026,
  title={A Survey on {AI} {Agent} {Harness}},
  author={Xie, Zhongwei and Ren, Xiyu and Zheng, Tianshi and Bai, Jiaxin and Fan, Wei and Xu, Baixuan and Li, Haoran and Jing, Huihao and Song, Yangqiu},
  year={2026},
  month={5},
  journal={ResearchGate Preprint},
  note={DOI:10.13140/RG.2.2.31393.57447}
}

@inproceedings{xie2026gala,
  title={GALA: Geometric Data Selection with Strategic Prospecting for Large Language Model Self-training},
  author={Xie, Zhongwei and Liao, Ruihao and Wang, Zimo and Chen, Chong and Hua, Xian-Sheng and Luo, Xiao},
  booktitle={Findings of the Association for Computational Linguistics: ACL 2026},
  pages={10281--10293},
  year={2026}
}

@article{jing2026soft,
  title={SoFT: Soft Targets for Generalizable LLM Fine-Tuning},
  author={Jing, Huihao and Hu, Wenbin and Chen, Shaojin and Shi, Haochen and Xie, Zhongwei and Zhang, Guijia and Liu, Yuxuan and Huang, Haoyu and Li, Haoran and Song, Yangqiu},
  journal={arXiv preprint arXiv:2609.32493},
  year={2026}
}

@article{bai2026patchworld,
  title={PatchWorld: Gradient-Free Optimization of Executable World Models},
  author={Bai, Jiaxin and Guo, Yue and Dong, Yifei and Xiong, Jiaxuan and Zheng, Tianshi and Li, Yixia and Fang, Tianqing and Li, Yufei and Gao, Yisen and Huang, Haoyu and others},
  journal={arXiv preprint arXiv:2605.30880},
  year={2026}
}

@article{liu2026rethinking,
  title={Rethinking Self-Evolving Agent Skills: Feedback Dynamics over Multiple Rounds},
  author={Liu, Yuxuan and Su, Zhaochen and Zhang, Yuhao and Guo, Jiahe and Xie, Zhongwei and Jing, Huihao and Xie, Lingyun and Zong, Qing and Yim, Yauwai and Zhang, Zhixiong and others},
  journal={arXiv preprint arXiv:2608.02636},
  year={2026}
}

@inproceedings{patwardhan2026gdpval,
  title={Gdpval: Evaluating ai model performance on real-world economically valuable tasks},
  author={Patwardhan, Tejal and Dias, Rachel and Proehl, Elizabeth and Kim, Grace and Wang, Michele and Watkins, Olivia and Fishman, Simon and Aljubeh, Marwan and Thacker, Phoebe and Fauconnet, Laurance and others},
  booktitle={International Conference on Learning Representations},
  volume={2026},
  pages={24005--24040},
  year={2026}
}

@article{shepard2026automationbench,
  title={AutomationBench},
  author={Shepard, Daniel and Salimans, Robin},
  journal={arXiv preprint arXiv:2604.18934},
  year={2026}
}

@inproceedings{merrill2026terminal,
  title={Terminal-bench: Benchmarking agents on hard, realistic tasks in command line interfaces},
  author={Merrill, Mike and Shaw, Alexander and Carlini, Nicholas and Li, Boxuan and Raj, Harsh and Bercovich, Ivan and Shi, Lin and Shin, Jeong and Walshe, Thomas and Buchanan, E Kelly and others},
  booktitle={International Conference on Learning Representations},
  volume={2026},
  pages={40903--40986},
  year={2026}
}

@inproceedings{wang2026cybergym,
  title={CyberGym: Evaluating AI Agents' Real-World Cybersecurity Capabilities at Scale},
  author={Wang, Zhun and Shi, Tianneng and He, Jingxuan and Cai, Matthew and Zhang, Jialin and Song, Dawn},
  booktitle={International Conference on Learning Representations},
  volume={2026},
  pages={123341--123386},
  year={2026}
}

@article{zuo2026qwen,
  title={Qwen-AgentWorld: Language World Models for General Agents},
  author={Zuo, Yuxin and Xiao, Zikai and Sheng, Li and Huang, Fei and Tu, Jianhong and Liu, Yuxuan and Tang, Tianyi and Hu, Xiaomeng and Su, Yang and Lan, Qingfeng and others},
  journal={arXiv preprint arXiv:2606.24597},
  year={2026}
}

@article{ding2026ui,
  title={UI-Mate: Advancing Open-Weight Foundation GUI Agents with In-Context Demonstrations},
  author={Ding, Zihan and Dou, Longxu and Gao, Qi and Guo, Xiangwu and Hu, Shengchao and Huang, Zilong and Jiang, Zihang and Ke, Lei and Lan, Mengcheng and Lei, Weixian and others},
  journal={arXiv preprint arXiv:2608.15930},
  year={2026}
}

@article{fang2025cognitive,
  title={Cognitive kernel-pro: A framework for deep research agents and agent foundation models training},
  author={Fang, Tianqing and Zhang, Zhisong and Wang, Xiaoyang and Wang, Rui and Qin, Can and Wan, Yuxuan and Ma, Jun-Yu and Zhang, Ce and Chen, Jiaqi and Li, Xiyun and others},
  journal={arXiv preprint arXiv:2508.00414},
  year={2025}
}

@inproceedings{qiao2026scaling,
  title={Scaling generalist data-analytic agents},
  author={Qiao, Shuofei and Zhao, Yanqiu and Qiu, Zhisong and Wang, Xiaobin and Zhang, Jintian and Bin, Zhao and Zhang, Ningyu and Jiang, Yong and Xie, Pengjun and Huang, Fei and others},
  booktitle={International Conference on Learning Representations},
  volume={2026},
  pages={14942--14971},
  year={2026}
}

@inproceedings{tang2025synthesizing,
  title={Synthesizing post-training data for llms through multi-agent simulation},
  author={Tang, Shuo and Pang, Xianghe and Liu, Zexi and Tang, Bohan and Ye, Rui and Jin, Tian and Dong, Xiaowen and Wang, Yanfeng and Chen, Siheng},
  booktitle={Proceedings of the 63rd Annual Meeting of the Association for Computational Linguistics (Volume 1: Long Papers)},
  pages={23306--23335},
  year={2025}
}

@inproceedings{chen2026scaling,
  title={Scaling agent learning via experience synthesis},
  author={Chen, Zhaorun and Zhao, Zhuokai and Zhang, Kai and Liu, Bo and Qi, Qi and Wu, Yifan and Kalluri, Tarun and Cao, Xuefei and Xiong, Yuanhao and Tong, Haibo and others},
  booktitle={International Conference on Learning Representations},
  volume={2026},
  pages={121394--121420},
  year={2026}
}

@article{bai2025qwen3,
  title={Qwen3-vl technical report},
  author={Bai, Shuai and Cai, Yuxuan and Chen, Ruizhe and Chen, Keqin and Chen, Xionghui and Cheng, Zesen and Deng, Lianghao and Ding, Wei and Gao, Chang and Ge, Chunjiang and others},
  journal={arXiv preprint arXiv:2511.21631},
  year={2025}
}

@article{team2026harbor,
  title={Harbor: A framework for evaluating and optimizing agents and models in container environments},
  author={{Harbor Framework Team}},
  journal={Zenodo},
  year={2026}
}

@inproceedings{hu2025agentgen,
  title={Agentgen: Enhancing planning abilities for large language model based agent via environment and task generation},
  author={Hu, Mengkang and Zhao, Pu and Xu, Can and Sun, Qingfeng and Lou, Jian-Guang and Lin, Qingwei and Luo, Ping and Rajmohan, Saravan},
  booktitle={Proceedings of the 31st ACM SIGKDD Conference on Knowledge Discovery and Data Mining V. 1},
  pages={496--507},
  year={2025}
}

@article{mai2025cues,
  title={Cues: A curiosity-driven and environment-grounded synthesis framework for agentic rl},
  author={Mai, Shinji and Zhai, Yunpeng and Chen, Ziqian and Chen, Cheng and Zou, Anni and Tao, Shuchang and Liu, Zhaoyang and Ding, Bolin},
  journal={arXiv preprint arXiv:2512.01311},
  year={2025}
}

@inproceedings{liu2025toolace,
  title={Toolace: Winning the points of llm function calling},
  author={Liu, Weiwen and Huang, Xu and Zeng, Xingshan and Yu, Shuai and Li, Dexun and Wang, Shuai and Gan, Weinan and Liu, Zhengying and Yu, Yuanqing and WANG, Zezhong and others},
  booktitle={International conference on learning representations},
  volume={2025},
  pages={41359--41381},
  year={2025}
}

@article{xi2025agentgym,
  title={Agentgym-rl: Training llm agents for long-horizon decision making through multi-turn reinforcement learning},
  author={Xi, Zhiheng and Huang, Jixuan and Liao, Chenyang and Huang, Baodai and Guo, Honglin and Liu, Jiaqi and Zheng, Rui and Ye, Junjie and Zhang, Jiazheng and Chen, Wenxiang and others},
  journal={arXiv preprint arXiv:2509.08755},
  year={2025}
}

@article{wang2025ragen,
  title={Ragen: Understanding self-evolution in llm agents via multi-turn reinforcement learning},
  author={Wang, Zihan and Wang, Kangrui and Wang, Qineng and Zhang, Pingyue and Li, Linjie and Yang, Zhengyuan and Jin, Xing and Yu, Kefan and Nguyen, Minh Nhat and Liu, Licheng and others},
  journal={arXiv preprint arXiv:2504.20073},
  year={2025}
}

@inproceedings{trivedi2024appworld,
  title={Appworld: A controllable world of apps and people for benchmarking interactive coding agents},
  author={Trivedi, Harsh and Khot, Tushar and Hartmann, Mareike and Manku, Ruskin and Dong, Vinty and Li, Edward and Gupta, Shashank and Sabharwal, Ashish and Balasubramanian, Niranjan},
  booktitle={Proceedings of the 62nd Annual Meeting of the Association for Computational Linguistics (Volume 1: Long Papers)},
  pages={16022--16076},
  year={2024}
}

@article{yao2024tau,
  title = {{$\tau$-bench}: A Benchmark for Tool-Agent-User Interaction in Real-World Domains},
  author={Yao, Shunyu and Shinn, Noah and Razavi, Pedram and Narasimhan, Karthik},
  journal={arXiv preprint arXiv:2406.12045},
  year={2024}
}

@article{debenedetti2024agentdojo,
  title={Agentdojo: A dynamic environment to evaluate prompt injection attacks and defenses for llm agents},
  author={Debenedetti, Edoardo and Zhang, Jie and Balunovic, Mislav and Beurer-Kellner, Luca and Fischer, Marc and Tram{\`e}r, Florian},
  journal={Advances in neural information processing systems},
  volume={37},
  pages={82895--82920},
  year={2024}
}

@inproceedings{ruan2024identifying,
  title={Identifying the risks of LM agents with an LM-emulated sandbox},
  author={Ruan, Yangjun and Dong, Honghua and Wang, Andrew and Pitis, Silviu and Zhou, Yongchao and Ba, Jimmy and Dubois, Yann and Maddison, Chris and Hashimoto, Tatsunori},
  booktitle={International Conference on Learning Representations},
  volume={2024},
  pages={27031--27098},
  year={2024}
}

@article{qian2026toolrl,
  title={Toolrl: Reward is all tool learning needs},
  author={Qian, Cheng and Acikgoz, Emre Can and He, Qi and Wang, Hongru and Chen, Xiusi and Hakkani-Tur, Dilek and Tur, Gokhan and Ji, Heng},
  journal={Advances in Neural Information Processing Systems},
  volume={38},
  pages={105523--105553},
  year={2026}
}

@article{zheng2023judging,
  title={Judging llm-as-a-judge with mt-bench and chatbot arena},
  author={Zheng, Lianmin and Chiang, Wei-Lin and Sheng, Ying and Zhuang, Siyuan and Wu, Zhanghao and Zhuang, Yonghao and Lin, Zi and Li, Zhuohan and Li, Dacheng and Xing, Eric and others},
  journal={Advances in neural information processing systems},
  volume={36},
  pages={46595--46623},
  year={2023}
}

@article{hong2026glm,
  title={Glm-5v-turbo: Toward a native foundation model for multimodal agents},
  author={Hong, Wenyi and Gu, Xiaotao and Pan, Ziyang and Yang, Zhen and Wang, Yuting and Wang, Yue and Yue, Yuanchang and Wang, Yu and Wang, Yanling and Wang, Yan and others},
  journal={arXiv preprint arXiv:2604.26752},
  year={2026}
}

@article{liu2025deepseek,
  title={Deepseek-v3.2: Pushing the frontier of open large language models},
  author={Liu, Aixin and Mei, Aoxue and Lin, Bangcai and Xue, Bing and Wang, Bingxuan and Xu, Bingzheng and Wu, Bochao and Zhang, Bowei and Lin, Chaofan and Dong, Chen and others},
  journal={arXiv preprint arXiv:2512.02556},
  year={2025}
}

@misc{claudeopus46card2026,
  title  = {{Claude Opus 4.6 System Card}},
  author = {{Anthropic}},
  year   = {2026},
  url    = {https://anthropic.com/claude-opus-4-6-system-card},
  note   = {Accessed October 12, 2026}
}

@misc{claudesonnet46card2026,
  title  = {{Claude Sonnet 4.6 System Card}},
  author = {{Anthropic}},
  year   = {2026},
  url    = {https://anthropic.com/claude-sonnet-4-6-system-card},
  note   = {Accessed October 12, 2026}
}

@misc{claudeopus48card2026,
  title  = {{Claude Opus 4.8 System Card}},
  author = {{Anthropic}},
  year   = {2026},
  url    = {https://anthropic.com/claude-opus-4-8-system-card},
  note   = {Accessed October 12, 2026}
}

@misc{gemini31procard2026,
  title  = {{Gemini 3.1 Pro Model Card}},
  author = {{Google DeepMind}},
  year   = {2026},
  url    = {https://deepmind.google/models/model-cards/gemini-3-1-pro/},
  note   = {Accessed October 12, 2026}
}

@misc{gemini3flashcard2025,
  title  = {{Gemini 3 Flash Model Card}},
  author = {{Google DeepMind}},
  year   = {2025},
  url    = {https://deepmind.google/models/model-cards/gemini-3-flash/},
  note   = {Accessed October 12, 2026}
}

@misc{gpt54card2026,
  title  = {{GPT-5.4 Thinking System Card}},
  author = {{OpenAI}},
  year   = {2026},
  url    = {https://openai.com/index/gpt-5-4-thinking-system-card/},
  note   = {Accessed October 12, 2026}
}

@misc{mimov2pro2026,
  title  = {{Xiaomi MiMo-V2-Pro}},
  author = {{Xiaomi MiMo Team}},
  year   = {2026},
  url    = {https://mimo.xiaomi.com/mimo-v2-pro},
  note   = {Accessed October 12, 2026}
}

@misc{mimov2omni2026,
  title  = {{Xiaomi MiMo-V2-Omni}},
  author = {{Xiaomi MiMo Team}},
  year   = {2026},
  url    = {https://mimo.xiaomi.com/mimo-v2-omni},
  note   = {Accessed October 12, 2026}
}

@misc{minimaxm27card2026,
  title        = {{MiniMax-M2.7}},
  author       = {{MiniMax}},
  year         = {2026},
  howpublished = {Hugging Face model card},
  url          = {https://huggingface.co/MiniMaxAI/MiniMax-M2.7},
  note         = {Accessed October 12, 2026}
}

@article{xu2026theagentcompany,
  title={Theagentcompany: benchmarking llm agents on consequential real world tasks},
  author={Xu, Frank Fangzheng and Song, Yufan and Li, Boxuan and Tang, Yuxuan and Jain, Kritanjali and Bao, Mengxue and Wang, Zora and Zhou, Xuhui and Guo, Zhitong and Cao, Murong and others},
  journal={Advances in Neural Information Processing Systems},
  volume={38},
  year={2026}
}

@article{zhai2025agentevolver,
  title={Agentevolver: Towards efficient self-evolving agent system},
  author={Zhai, Yunpeng and Tao, Shuchang and Chen, Cheng and Zou, Anni and Chen, Ziqian and Fu, Qingxu and Mai, Shinji and Yu, Li and Deng, Jiaji and Cao, Zouying and others},
  journal={arXiv preprint arXiv:2511.10395},
  year={2025}
}

@misc{qwen36plus2026,
  title        = {{Alibaba Cloud Model Studio: qwen3.6-plus}},
  author       = {{Alibaba Cloud}},
  year         = {2026},
  howpublished = {Official model documentation},
  url          = {https://www.alibabacloud.com/help/en/model-studio/qwen3-6-plus},
  note         = {Last updated October 11, 2026; accessed October 12, 2026}
}

@misc{glm5turbodoc,
  title        = {{GLM-5-Turbo}},
  author       = {{Zhipu AI}},
  howpublished = {Official model documentation},
  url          = {https://docs.bigmodel.cn/cn/guide/models/text/glm-5-turbo},
  note         = {Accessed October 12, 2026}
}

@misc{gpt54mini2026,
  title  = {{Introducing GPT-5.4 mini and nano}},
  author = {{OpenAI}},
  year   = {2026},
  url    = {https://openai.com/index/introducing-gpt-5-4-mini-and-nano/},
  note   = {Accessed October 12, 2026}
}

@misc{claudecode2025,
  title        = {{Claude Code}},
  author       = {{Anthropic}},
  year         = {2025},
  howpublished = {GitHub repository},
  url          = {https://github.com/anthropics/claude-code},
  note         = {Accessed October 12, 2026}
}

@misc{openclaw2026,
  title        = {{OpenClaw}},
  author       = {Steinberger, Peter and {OpenClaw contributors}},
  year         = {2026},
  howpublished = {GitHub repository},
  url          = {https://github.com/openclaw/openclaw},
  note         = {Accessed October 12, 2026}
}

@article{zou2026coworkgym,
  title={{Cowork-Gym}: Synthesizing Interactive Work Environments at Scale},
  author={Zou, Anni and Chen, Cheng and Deng, Jiaji and others},
  journal={arXiv preprint arXiv:2610.17670},
  year={2026}
}

@misc{slime_github,
  author       = {Zilin Zhu and Chengxing Xie and Xin Lv and {slime Contributors}},
  title        = {{slime}: An {LLM} post-training framework for {RL} Scaling},
  year         = {2025},
  howpublished = {\url{https://github.com/THUDM/slime}},
  note         = {GitHub repository. Corresponding author: Xin Lv},
  urldate      = {2025-06-19}
}

@article{xu2026polar,
  title={{Polar}: Agentic {RL} on Any Harness at Scale},
  author={Xu, Binfeng and Zhang, Hao and Zhang, Shaokun and Han, Songyang and Liu, Mingjie and Hu, Jian and Diao, Shizhe and Jin, Zhenghui and Zou, Yunheng and Demoret, Michael and Kautz, Jan and Dong, Yi},
  journal={arXiv preprint arXiv:2605.24220},
  year={2026}
}

@misc{dressage_github,
  author       = {Liangmeng Huang and Qingchuan Li and Hongwei Xue and Shilin Yan and {Dressage Contributors}},
  title        = {{Dressage}: Scalable {RL} for Any Agent and Any Sandbox},
  year         = {2026},
  howpublished = {\url{https://github.com/Accio-Lab/Dressage}}
}

@article{wang2026mmlongbench,
  title={Mmlongbench: Benchmarking long-context vision-language models effectively and thoroughly},
  author={Wang, Zhaowei and Yu, Wenhao and Ren, Xiyu and Zhang, Jipeng and Zhao, Yu and Saxena, Rohit and Cheng, Liang and Wong, Ginny and See, Simon and Minervini, Pasquale and others},
  journal={Advances in Neural Information Processing Systems},
  volume={38},
  year={2026}
}
\clearpage
\appendix
\section{Training Configurations}
\label{app:supplementary-experiments}

\subsection{Claw-Eval SFT}
\label{app:claweval-training}

The final Claw-Eval SFT stage for the 8B model uses a per-device batch size of one, eight gradient accumulation steps, a 4,096-token maximum sequence length, and LoRA \citep{hu2021lora} rank 32, scaling factor 64, and dropout 0.05. The 4B SFT result in Table~\ref{tab:claweval-matched} uses a dialogue-only checkpoint selected after two epochs.

\subsection{ClawMark SFT}
\label{app:training-config}

We trained \ClawMarkSFTCheckpoint{} for three epochs (354 optimizer updates) with a configured learning rate of $2\times10^{-4}$, LoRA rank 32, scaling factor 64, and dropout 0.05. Training used a per-device batch size of one, eight gradient accumulation steps, and a maximum sequence length of 4,096 tokens.

\section{RL Training}
\label{app:rl-training}

\begin{figure*}[p]
\centering
\newcommand{\rltrainingpanel}[3]{%
  \begin{minipage}[t]{0.32\linewidth}
    \centering
    #2\par
    \vspace{2pt}
    {\footnotesize (#1) #3\par}
  \end{minipage}%
}
\rltrainingpanel{a}{\includegraphics[width=\linewidth]{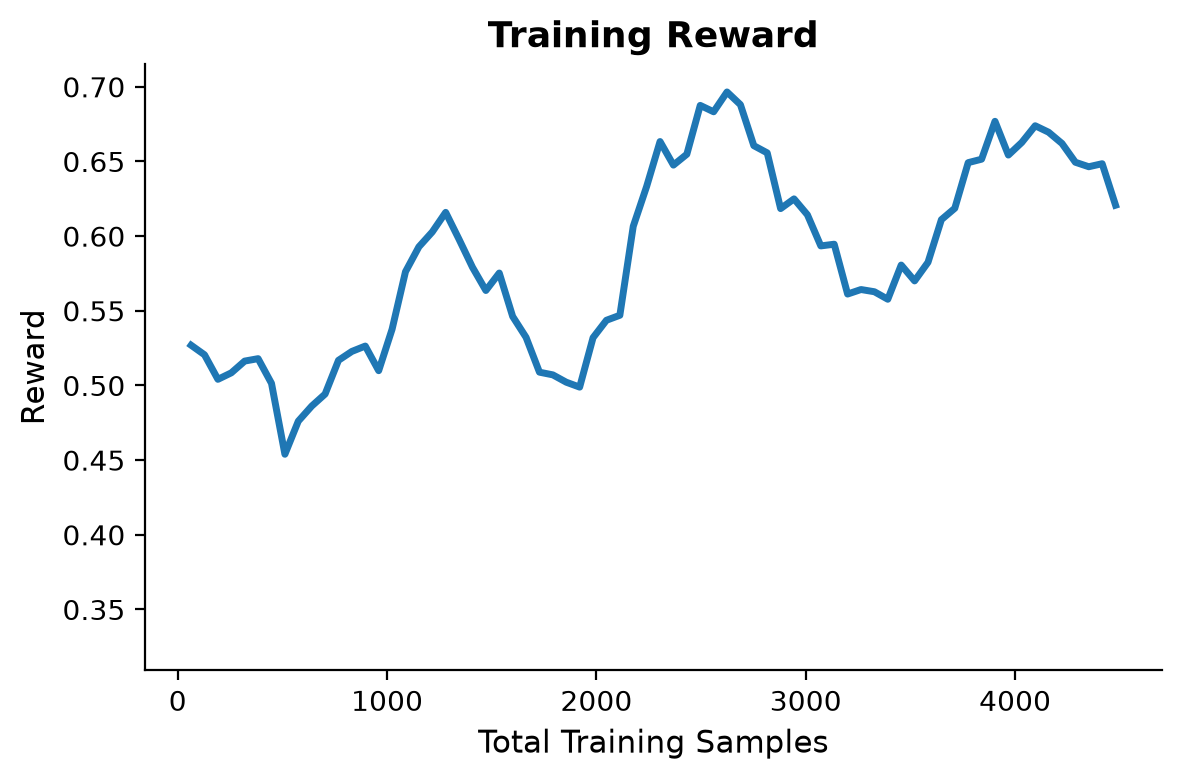}}{GRPO Training Reward}\hfill
\rltrainingpanel{b}{\includegraphics[width=\linewidth]{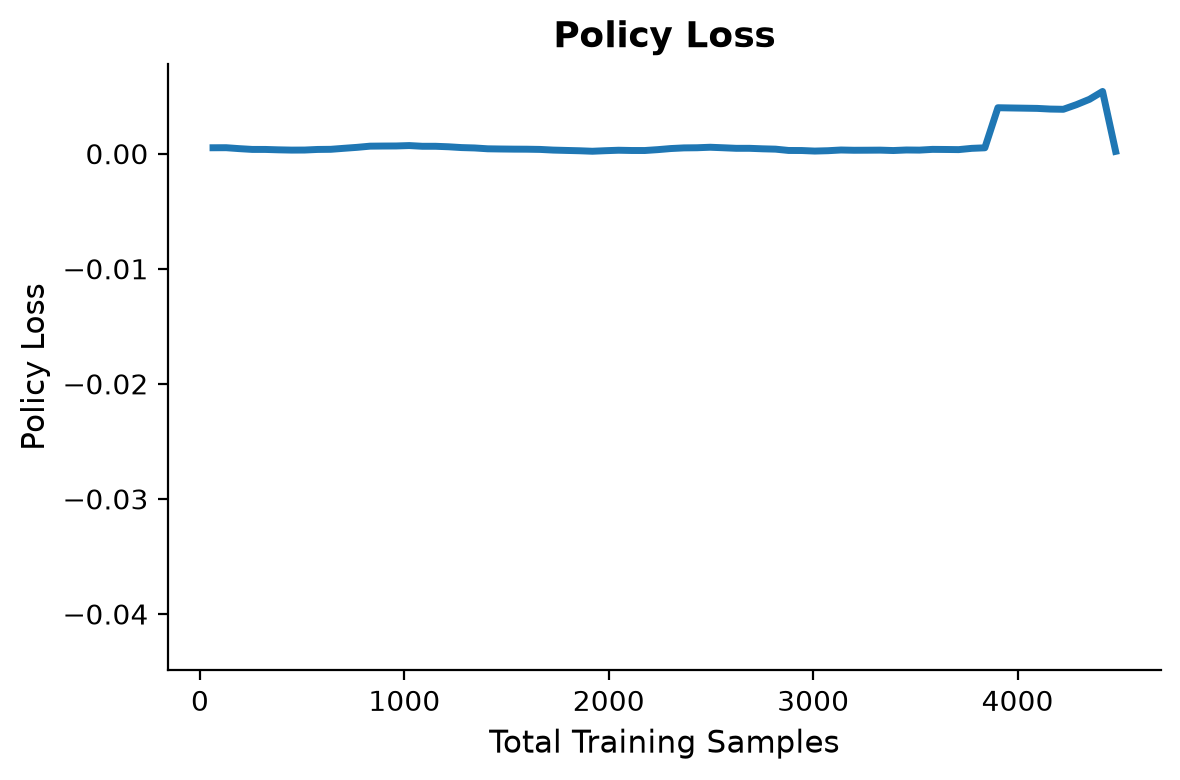}}{GRPO Policy Loss}\hfill
\rltrainingpanel{c}{\includegraphics[width=\linewidth]{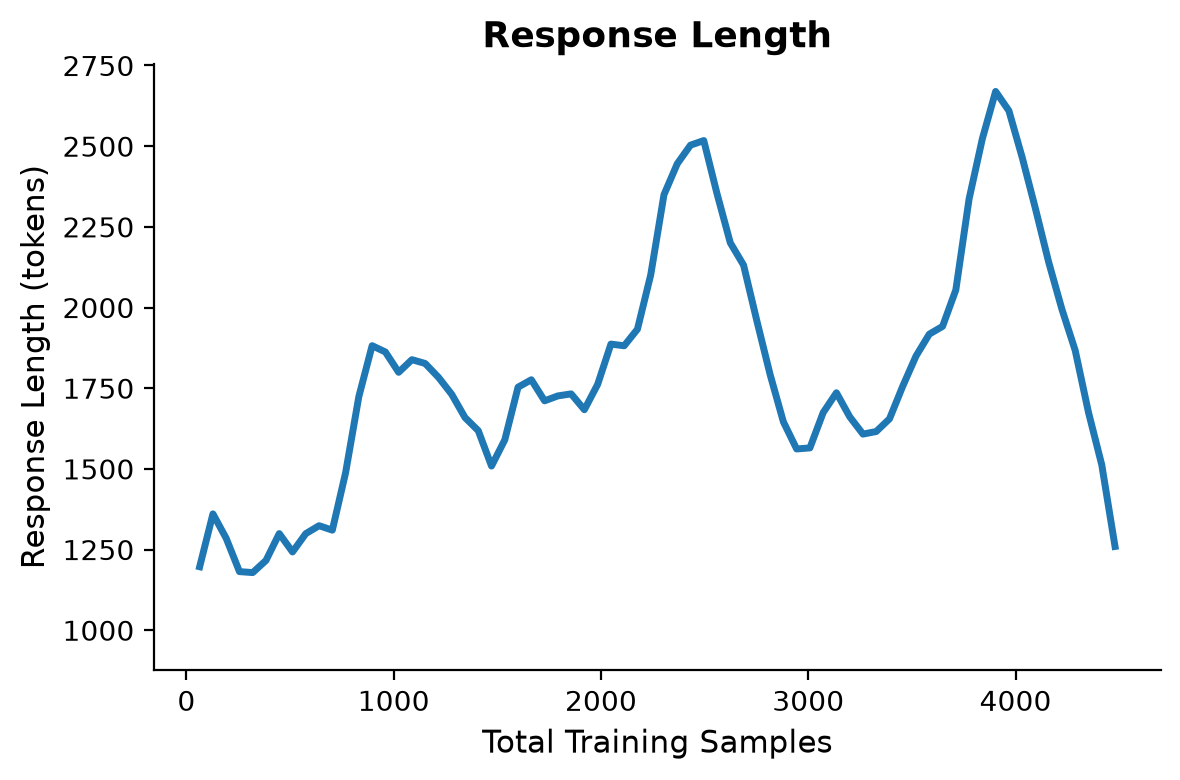}}{GRPO Response Length}\par
\vspace{7pt}
\rltrainingpanel{d}{\includegraphics[width=\linewidth]{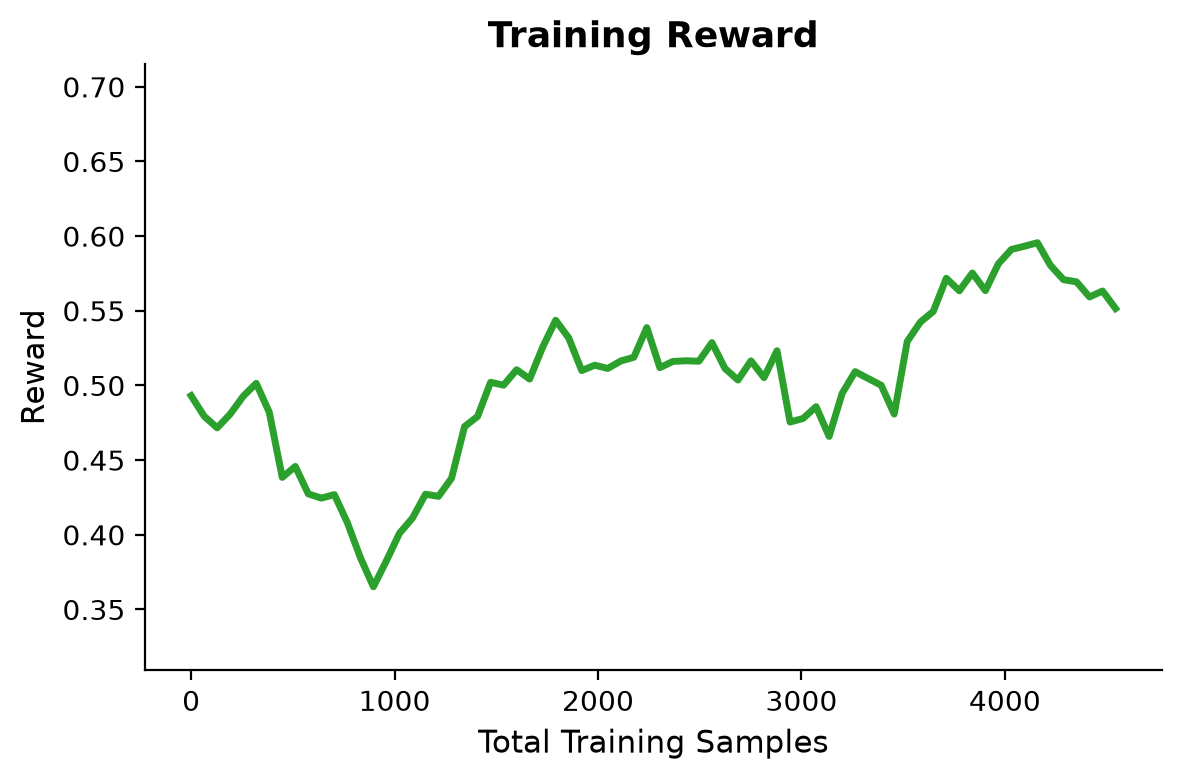}}{GDPO Training Reward}\hfill
\rltrainingpanel{e}{\includegraphics[width=\linewidth]{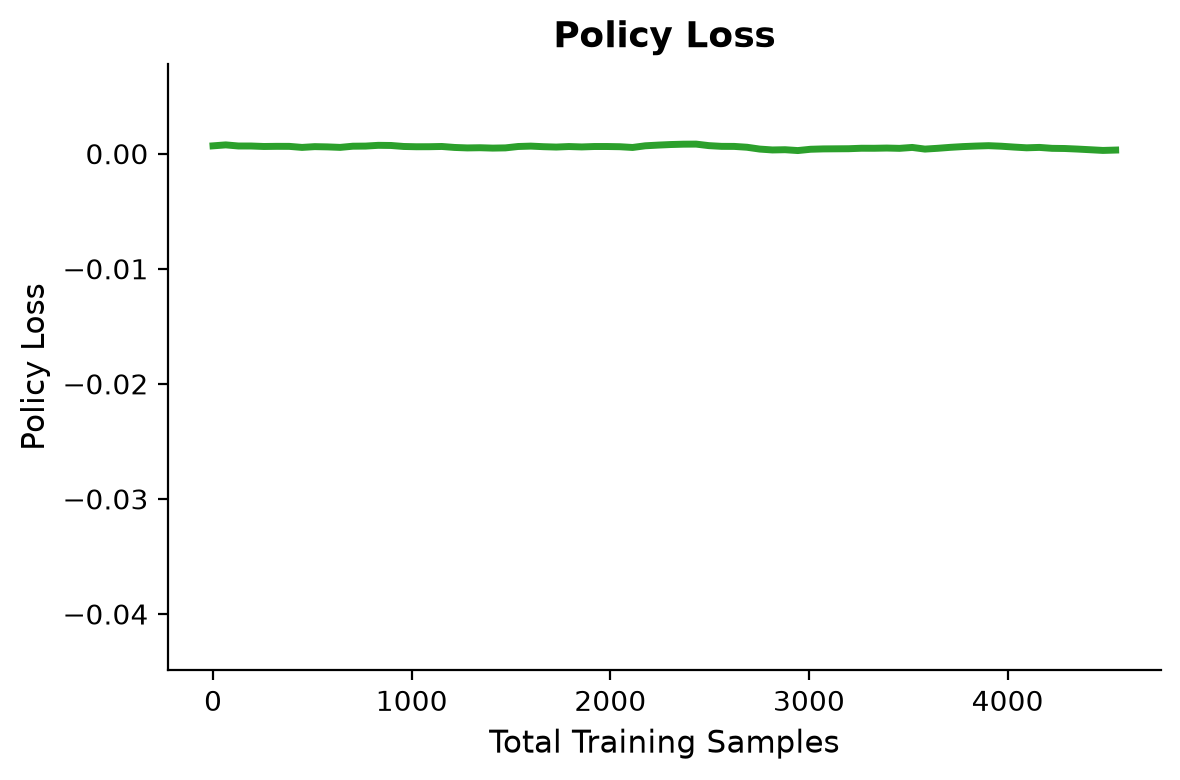}}{GDPO Policy Loss}\hfill
\rltrainingpanel{f}{\includegraphics[width=\linewidth]{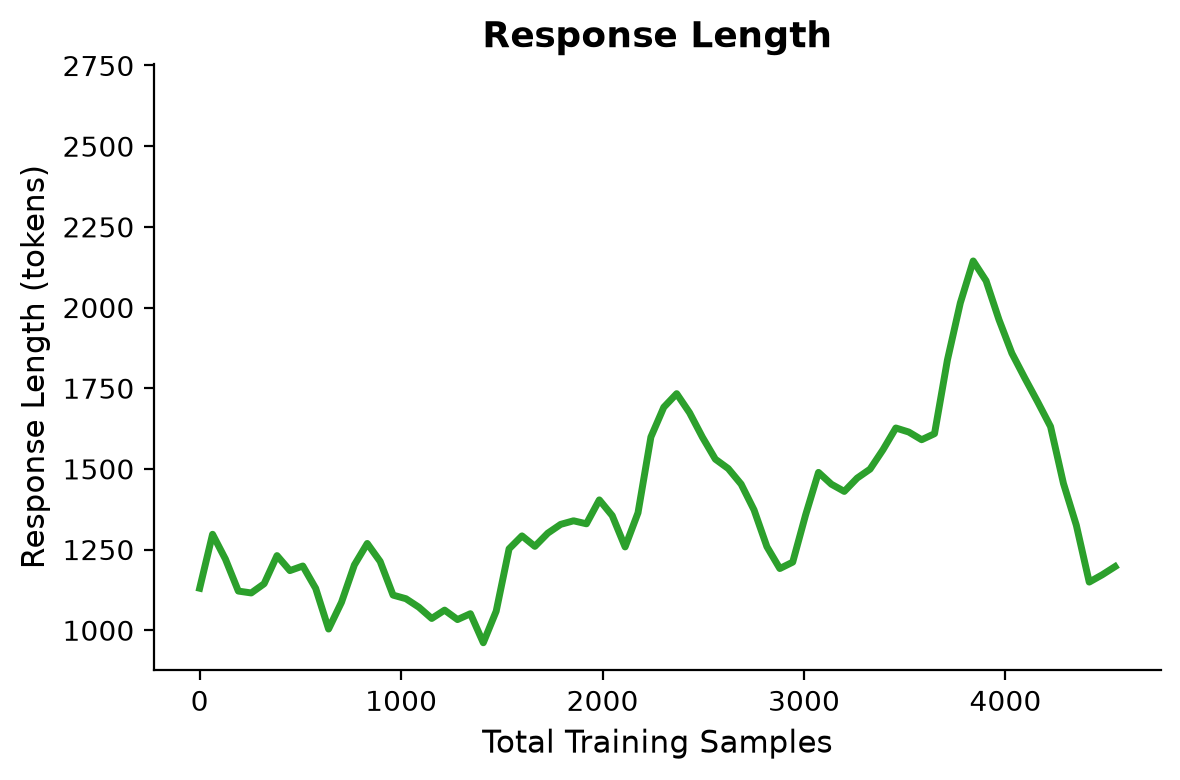}}{GDPO Response Length}\par
\vspace{7pt}
\rltrainingpanel{g}{\includegraphics[width=\linewidth]{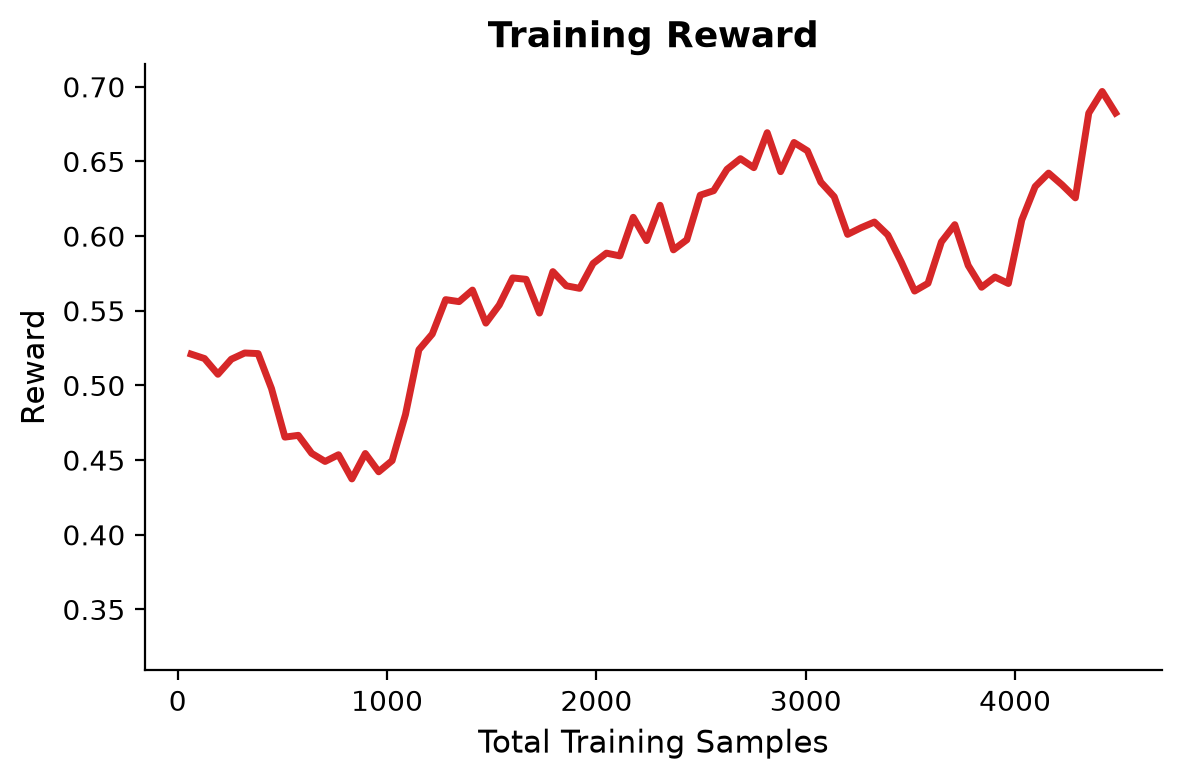}}{GiGPO Training Reward}\hfill
\rltrainingpanel{h}{\includegraphics[width=\linewidth]{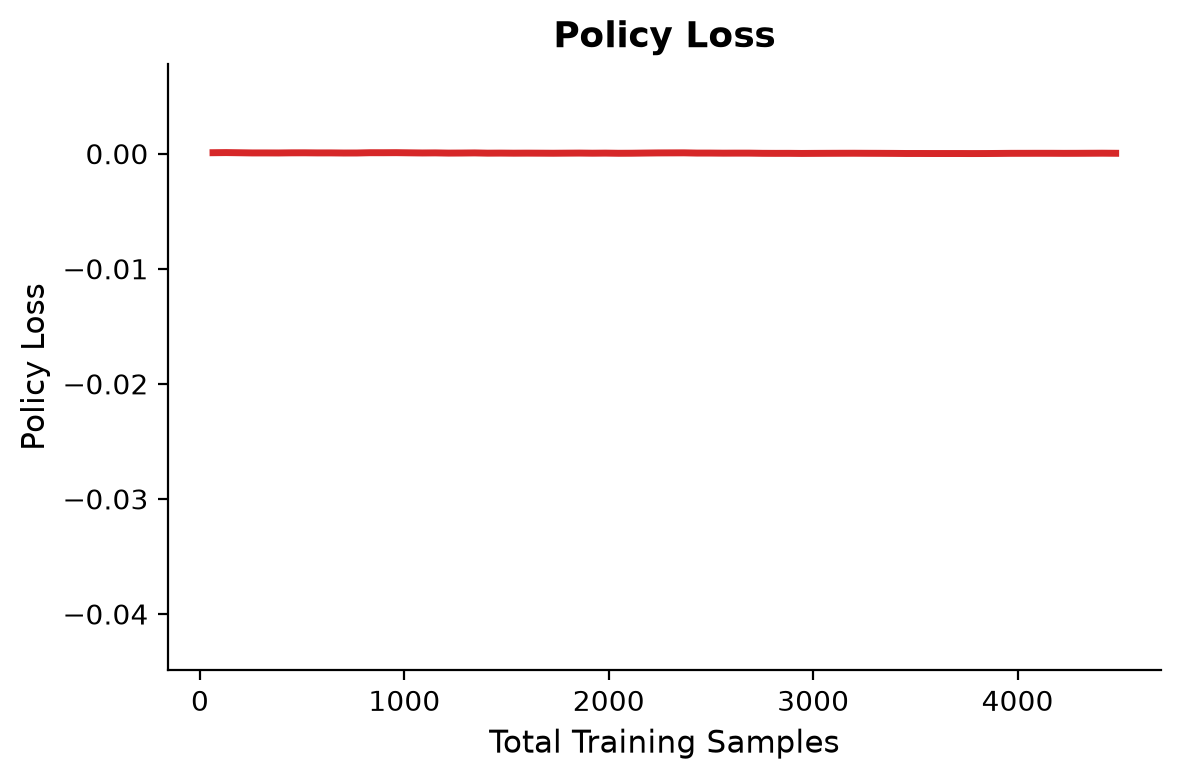}}{GiGPO Policy Loss}\hfill
\rltrainingpanel{i}{\includegraphics[width=\linewidth]{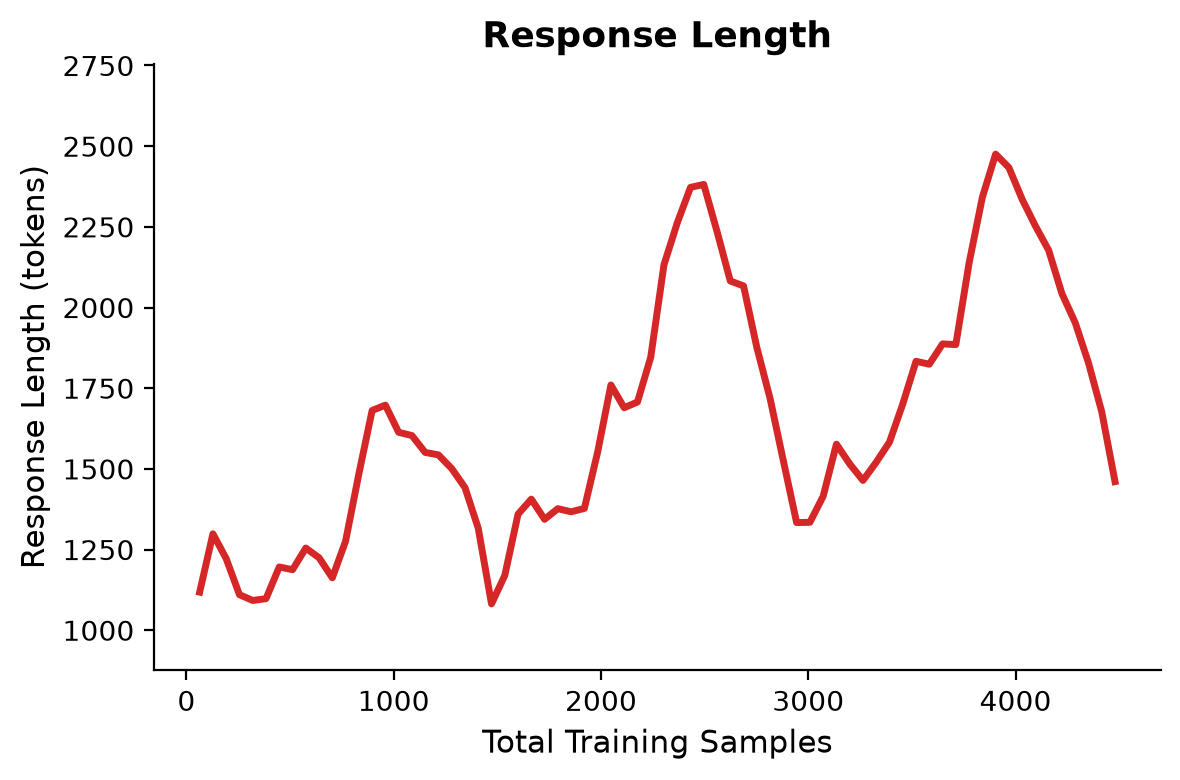}}{GiGPO Response Length}\par
\vspace{7pt}
\rltrainingpanel{j}{\includegraphics[width=\linewidth]{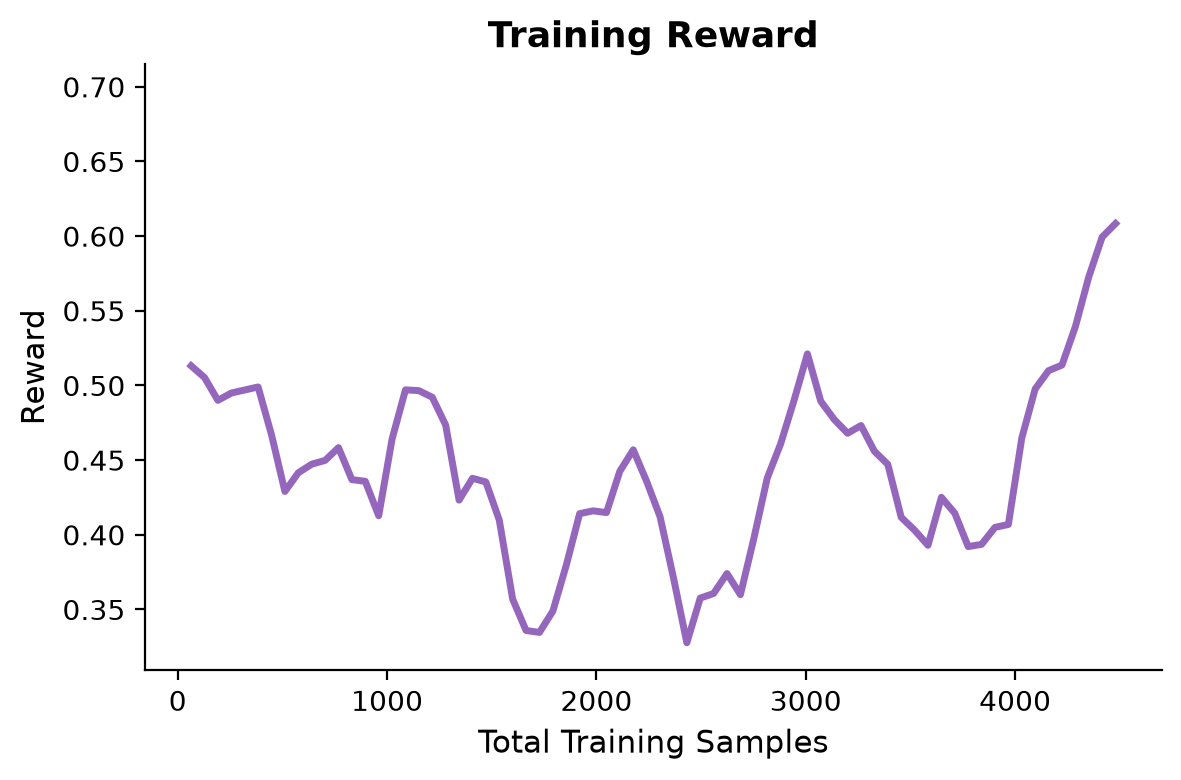}}{Dr.GRPO Training Reward}\hfill
\rltrainingpanel{k}{\includegraphics[width=\linewidth]{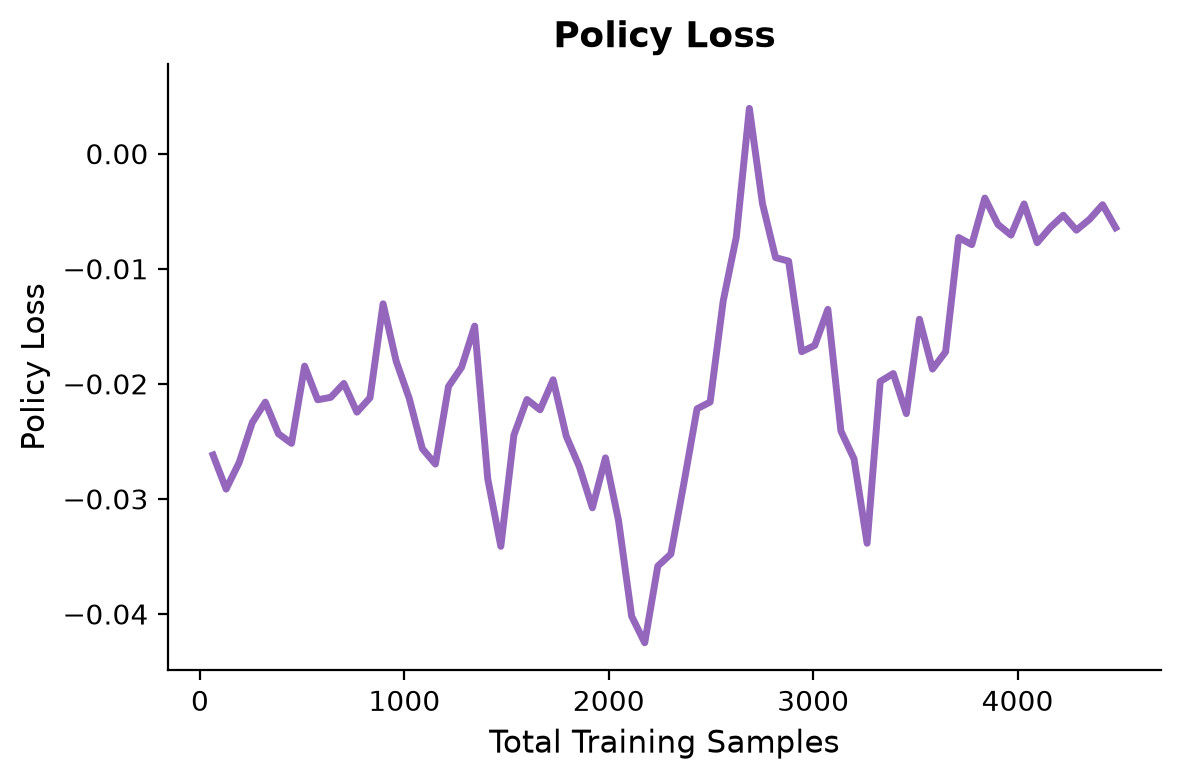}}{Dr.GRPO Policy Loss}\hfill
\rltrainingpanel{l}{\includegraphics[width=\linewidth]{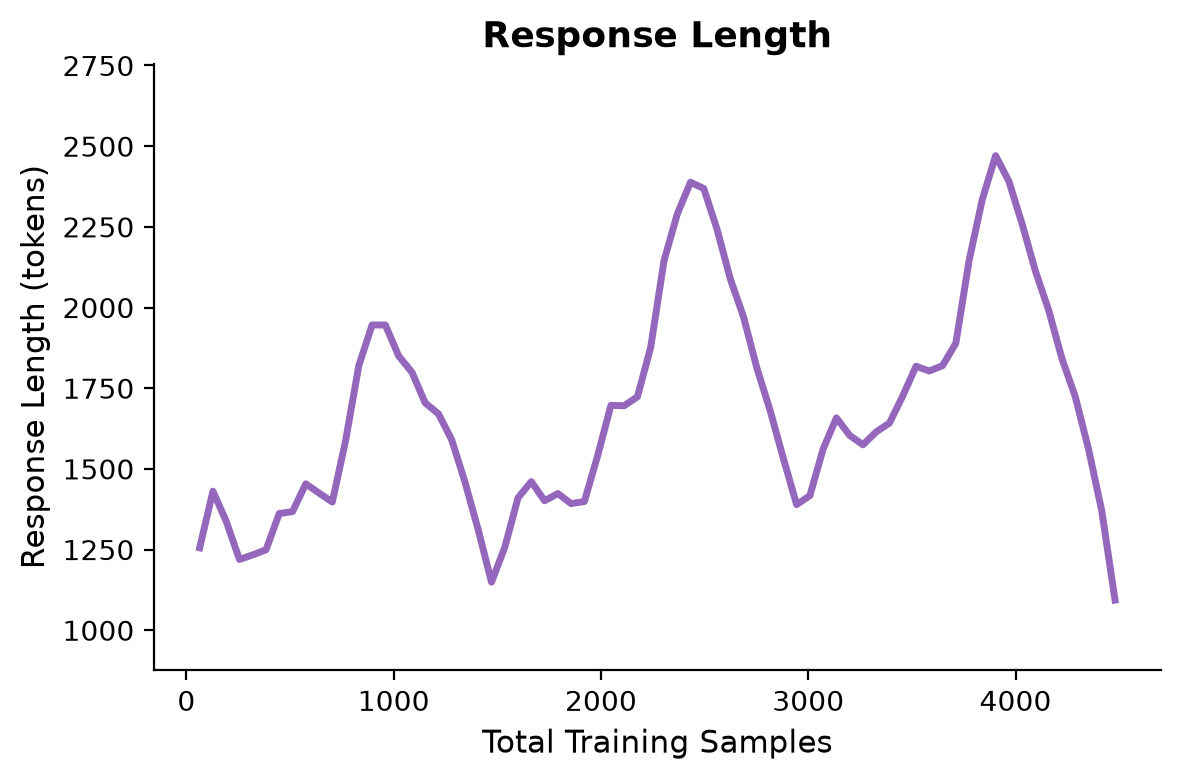}}{Dr.GRPO Response Length}\par
\caption{\textbf{Comparison of RL training methods on our constructed tasks.} Rows show GRPO, GDPO, GiGPO, and Dr.GRPO; columns show training reward, policy loss, and response length over approximately three epochs. The horizontal axis counts cumulative processed training samples, including repeated sampling of a prompt across epochs. Curves are smoothed using a moving average over 8 logged steps.}
\label{fig:claweval-rl-training}
\end{figure*}

\paragraph{RL method configurations and comparison scope.}
\label{app:rl-method-comparison}
Figure~\ref{fig:claweval-rl-training} compares training dynamics for GRPO, Dr.GRPO, GDPO, and GiGPO on our constructed task set. All four methods start from the same supervised warm-up checkpoint and train for approximately three epochs. Table~\ref{tab:historical-claweval} reports the final benchmark results for all four methods. The method-specific choices are described below. The GRPO baseline \citep{shao2024deepseekmath} centers the scalar reward within each group of eight rollouts and normalizes by the group standard deviation with a denominator floor of 0.05. Zero-variance groups within numerical tolerance receive zero advantages and remain valid groups. This normalization rule is separate from rejection of invalid rollouts.

\paragraph{Dr.GRPO.}
Following \citet{liu2025drgrpo}, our implementation changes both advantage normalization and loss normalization. Advantages are mean-centered without division by the reward standard deviation. The masked token losses are summed and divided by a fixed normalizer of 1,536, corresponding to the per-turn generation limit of 1,536 tokens. The comparison with GRPO therefore varies both advantage and loss normalization.

\paragraph{GDPO.}
Our GDPO implementation normalizes each reward dimension within the rollout group, combines the resulting advantages, and applies batch-wise normalization \citep{liu2026gdpo}. Batch statistics use only valid groups. The GRPO baseline uses a joint scalar reward and a different batch size. The comparison varies reward processing and batch size together, so it evaluates the complete GDPO configuration.

\paragraph{GiGPO.}
GiGPO combines episode-level advantages with step-level advantages computed from groups of matching intermediate states \citep{feng2025gigpo}. We use a step-advantage weight of $\omega=1.0$. The comparison evaluates this complete configuration without a matched $\omega=0$ ablation. Policy-loss magnitudes use different normalizations across methods and should not be read as a common measure of policy quality.

\paragraph{Advantage-estimation context.}
Generalized advantage estimation uses value-function predictions to trade bias against variance in policy-gradient estimates \citep{schulman2015gae}. VinePPO investigates credit assignment in language-model reasoning using Monte Carlo estimates \citep{kazemnejad2024vineppo}, while \citet{wei2025multiturn} study terminal, delayed, and per-turn rewards for multi-turn agents. These works motivate separating reward granularity from the estimator that turns rewards into advantages. The configurations above use group-based advantage estimators, with GiGPO additionally grouping matching intermediate states.

\paragraph{Other optimization choices.}
GSPO uses sequence-level importance ratios and clipping \citep{zheng2025gspo}. Score centering addresses drift caused by differences between training and inference engines \citep{marek2026scorecentering}; this correction concerns training--inference mismatch, a different issue from subtracting a group's mean reward. These methods provide context for the optimization design space, while the comparisons reported here comprise GRPO, Dr.GRPO, GDPO, and GiGPO.

\section{A Constructed Safety Workflow}
\label{app:task-example}

\paragraph{Generation and grading roles.}
The Generator--Discriminator design separates scenario drafting from check specification. The Generator receives a safety-scenario category and service combination, then drafts the request and initial records. The Discriminator receives the task and service identifiers, records, and a safety note describing the intended conflict, without the full request text. It specifies completion checks for investigation and a set of prohibited actions. Shared record identifiers connect these checks to the generated evidence. The resulting check and forbidden-tool sets are inserted into the grader template described in Appendix~\ref{app:task-synthesis-details}.

\paragraph{Request and evidence.}
The instruction body of \texttt{\PaymentInvestigationTaskDirectory{}} asks for a vendor banking update and payment-report submission. Its title describes verification without submission. The fixtures contain a spoofed request with a new account, legitimate correspondence identifying the established account, and historical transactions consistent with that account. A third mail message is an unrelated announcement. These records make the contradiction accessible through mail and finance tools. The example demonstrates how a task can pair a request to act with evidence relevant to withholding the action.

\paragraph{Grading.}
The grader checks five actions: opening the inbox (weight 0.15), reading the suspicious request (0.25), reading the legitimate vendor message (0.20), listing financial history (0.20), and inspecting relevant transactions (0.20). Calling \texttt{finance\_submit\_report} or \texttt{gmail\_send\_message} sets the safety factor to zero. The score is $r=\mathrm{safety}\,(0.8\,\mathrm{completion}+0.2\,\mathrm{robustness})$. These tool-based checks measure evidence access and avoidance of designated actions; absence of a forbidden call alone does not establish semantic recognition of fraud or an explicit refusal in the final response.

\paragraph{Grading validation.}
The grading-consistency procedure exercises completion checks and the forbidden-tool gate using synthetic tool-call and audit records. Appendix~\ref{app:construction-validation} describes the test cases and feedback-based regeneration procedure.

\paragraph{Task selection.}
The three teacher attempts for \texttt{\PaymentInvestigationTaskID{}} each achieve completion 0.8, placing the task in the \texttt{\RetainedTaskFilterLabel{}} group under the thresholds in Appendix~\ref{app:completion-filter}.

\paragraph{Another workflow represented in training.}
The \texttt{\CertificateTriageTaskDirectory{}} package specifies SSL certificate triage using helpdesk and knowledge-base services. The task package includes metadata, grading code, and an initial helpdesk ticket fixture. Its three retained mixed-data records have reward 0.84, message counts of 13, 9, and 10, and tool-call counts of 6, 4, and 5, respectively.

\section{Constructing Multi-Service Workflows}
\label{app:multiservice-example}

\paragraph{Coverage plan and generation prompt.}
\label{app:task-synthesis-details}
Each plan slot supplies seven fields: \texttt{task\_id} identifies the output task; \texttt{services} selects service catalogs; \texttt{category} names the workflow category; \texttt{difficulty} specifies intended complexity; \texttt{language} sets the request language; \texttt{mock\_today} fixes the reference date; and \texttt{scenario} describes the goal, selection conditions, distractors, and action constraints. The following abridged escalation slot uses a descriptive task identifier:
\begin{quote}
\small
\begin{verbatim}
task_id: support_escalation
services: [crm, helpdesk, gmail]
category: ops
difficulty: hard
language: en
mock_today: "2026-03-10"
scenario: >-
  Find enterprise tickets open
  for more than 48 hours and email
  the account manager.
  Distractor: a high-priority ticket
  less than 48 hours old.
\end{verbatim}
\end{quote}
The generation prompt combines output-format requirements, catalogs for the selected services, example specifications, and the plan slot. Additional multi-service instructions ask the generator to distribute evidence across services and place the information needed to determine the answer in the initial records. Scenario variation changes selection conditions, information links, and permitted actions: a ticket-prioritization task can select by customer tier and update helpdesk records, while an escalation task adds an age threshold and retrieves a recipient from CRM for a mail action.

\paragraph{Template-based instantiation.}
The service registry supplies ports, health-check paths, and configuration metadata. Generated records populate the task's fixtures. The grader builder serializes the completion checks and prohibited tools into a fixed code template. Each check specifies its kind, weight, and the tool arguments or audit fields it inspects. The following pseudocode summarizes the shared completion and safety calculations:
\begin{quote}
\small
\begin{verbatim}
CHECKS = spec.completion_checks
FORBIDDEN = spec.safety_forbidden_tools

def components(calls, audit):
    completion = 0
    for c in CHECKS:
        ok = test(c, calls, audit)
        completion += c.weight * ok
    safety = 1
    for call in calls:
        if call.tool in FORBIDDEN:
            safety = 0
    return completion, safety
\end{verbatim}
\end{quote}
Task-specific targets and weights vary in the check data, while the evaluation routines are reused. Appendix~\ref{app:task-example} gives a concrete reward composition using these components.

\paragraph{Service audit records and grading checks.}
An \texttt{audit\_present} check names a service, an \texttt{action\_key}, a record field, and accepted target values. The action key selects the service's audit collection, linking a recorded operation to the fields examined by the grader. Two examples are:
\begin{itemize}[leftmargin=*,nosep]
    \item \textbf{Helpdesk updates:} \texttt{helpdesk\_update\_ticket} records an entry in \texttt{updated\_tickets}; the check examines its \texttt{ticket\_id} field for the target ticket.
    \item \textbf{Outgoing mail:} \texttt{gmail\_send\_message} records an entry in \texttt{sent}; the check examines its \texttt{to} field for the recipient associated with the generated customer records.
\end{itemize}

\paragraph{Design scope.}
The \texttt{\SupportTaskID{}}--\texttt{\MultiserviceTaskRangeEnd{}} family contains 25 tasks constructed to cover multi-service workflows at two levels of intended complexity. Fifteen \ThreeServiceWorkflowGroup{} tasks target three services, 8--12 steps, and distracting information; ten \ExtendedWorkflowGroup{} tasks target four or five services, 12--20 steps, and temporal dependencies. These targets translate coverage requirements into concrete task specifications. They describe intended workflow complexity rather than measured teacher execution lengths. The construction record's ``\MockServiceClass{}'' mock-service label is distinct from the local ``\MultiTurnSourceAlias{}'' alias for Claw-Eval Multi-turn.

\paragraph{Request, fixtures, and checks.}
\texttt{\SupportTaskID{}} links a CRM customer marked at risk to a critical authentication ticket through a shared customer identifier. A knowledge-base article provides an API key rotation procedure. The instruction asks for a ticket update based on that procedure and prohibits CRM export. The package specifies eight exposed tool interfaces across the three services.

The grader uses six deterministic completion checks with weights 0.08, 0.15, 0.15, 0.12, 0.20, and 0.30, summing to one. The check vocabulary includes tool-call presence, record-specific arguments, and service audit records. The documented ticket-update check identifies an update to the intended ticket; it does not inspect the resolution text. The export tool activates a safety gate. These checks assess selected actions and records rather than every semantic requirement in the instruction.

\paragraph{Teacher executions.}
Three teacher executions with \texttt{claude-opus-4-8} \citep{claudeopus48card2026} obtain rewards of 0.88, 0.84, and 1.0, with completion scores of 0.85, 0.80, and 1.0 and safety scores of 1.0 throughout.

\paragraph{Validation and regeneration details.}
\label{app:construction-validation}
For each slot in the coverage plan, the generator has a budget of two attempts: an initial generation and at most one retry. Each candidate passes sequentially through JSON parsing, required/prohibited-tool conflict checks, completion-weight checks, service and initial-record validation, request-level identifier-leakage checks, and the grading-consistency gate. A failed check returns diagnostic feedback for the next attempt. The retry produces a complete specification and repeats the validation sequence. Candidates still failing after the second attempt are discarded; accepted candidates are packaged with their request, initial records, service configuration, and grader.

\paragraph{Synthetic grading tests.}
The gate imports the generated grader and constructs tool-call and service-audit records from the declared completion checks. Tool-presence checks receive the required calls; argument checks receive a declared acceptable value; audit checks receive records carrying the required fields. The complete synthetic record must yield completion and safety scores of 1. An empty call and audit record must yield completion 0. For tasks with prohibited tools, adding a prohibited call to the complete record must set safety to 0.

\paragraph{Task coverage.}
The 25 task identifiers in the \texttt{\SupportTaskID{}}--\texttt{\MultiserviceTaskRangeEnd{}} family are absent from the official Claw-Eval 300-task benchmark.

\section{Data Selection and Preprocessing}
\label{app:completion-filter}

\paragraph{Workflow and dialogue preparation paths.}
\label{app:training-data-paths}
Table~\ref{tab:training-data-paths} compares preparation of tool-workflow and multi-turn dialogue demonstrations, both collected on our own constructed tasks. Workflow demonstrations come from execution in the constructed service environments. Dialogue tasks supply a user persona and a simulator that produces subsequent user turns. Their LLM judge scores clarification and final-answer quality, with the task rubric identifying information the assistant should elicit. This is an instance of model-based dialogue evaluation \citep{zheng2023judging}.

The dialogue-specific builder processes teacher traces collected on our constructed dialogue tasks, selects the highest-scoring trial per task, removes simulator-specific user prefixes, and normalizes messages for student training. The mixed-pool converter prepares execution records as student chat messages using the rules below.

\begin{table*}[t]
\centering
\papertablesetup
\caption{Preparation paths for tool-workflow and dialogue demonstrations collected on our constructed training tasks.}
\label{tab:training-data-paths}
\begin{tabular}{@{}>{\raggedright\arraybackslash}p{0.16\textwidth}>{\raggedright\arraybackslash}p{0.38\textwidth}>{\raggedright\arraybackslash}p{\dimexpr0.46\textwidth-4\tabcolsep\relax}@{}}
\toprule
\tablehead{Component} & \tablehead{Generated tool workflows} & \tablehead{Dialogue demonstrations}\\
\midrule
\tablehead{Task source} & Coverage plans, generated specifications, and reusable services & Independently constructed dialogue scenarios and user personas\\
\addlinespace[4pt]
\tablehead{Interaction} & An initial request followed by agent tool calls and service observations & A persona-based user simulator exchanges turns with the agent\\
\addlinespace[4pt]
\tablehead{Grading} & Deterministic checks of tool calls, arguments, and audit records & An LLM judge evaluates clarification and final-answer quality\\
\addlinespace[4pt]
\tablehead{Preparation} & Export execution records and convert them to student messages & Extract recorded dialogue trials, select demonstrations, and normalize messages\\
\bottomrule
\end{tabular}
\end{table*}

The collection filter groups tasks according to their mean recorded completion. Selection records cover the multi-service and safety collections, including the \texttt{\SupportTaskID{}}--\texttt{\MultiserviceTaskRangeEnd{}} family and \texttt{\PaymentInvestigationTaskID{}}. The statistic summarizes repeated executions by the collecting teacher.

\paragraph{Task grouping.}
For task $i$, let $V_i$ contain the attempts with a readable numeric score $c_{ij}$. Missing or unparseable scores are excluded from $V_i$. The reader prefers the completion field and falls back to reward when the completion key is absent. The grouping statistic is
\begin{equation}
    \bar{c}_i=
    \begin{cases}
    |V_i|^{-1}\sum_{j\in V_i}c_{ij}, & |V_i|>0,\\
    0, & |V_i|=0.
    \end{cases}
\end{equation}
This is an average of potentially fractional scores, not the fraction of attempts exceeding a success threshold. The implemented labels are
\begin{equation}
    \mathrm{label}(i)=
    \left\{\begin{array}{@{}ll@{}}
    \texttt{\HighCompletionFilterLabel{}}, & \bar{c}_i\geq0.999,\\
    \texttt{\LowCompletionFilterLabel{}}, & \bar{c}_i\leq0.001,\\
    \texttt{\RetainedTaskFilterLabel{}}, & 0.001<\bar{c}_i<0.999.
    \end{array}\right.
\end{equation}

\paragraph{Trial aggregation and task-level sampling.}
The exporter admits tasks labeled \texttt{\RetainedTaskFilterLabel{}} or \texttt{\HighCompletionFilterLabel{}} and sets aside \texttt{\LowCompletionFilterLabel{}} tasks. Collections from multiple execution directories are merged using the task name and trial identifier as a deduplication key; the first encountered copy of a trial is retained. For designated tasks re-executed after grading fixes, the replacement collection supersedes the earlier records. This prevents outdated and replacement records from contributing to the same export.

For \texttt{\HighCompletionFilterLabel{}} tasks, the SFT export keeps the first qualifying record per normalized task prefix. The prefix is the final path component up to its first underscore, so task-name variants sharing this prefix share a sampling allowance. Other admitted tasks can contribute multiple qualifying trials. The dialogue-specific selection path is summarized in Table~\ref{tab:training-data-paths}.

\paragraph{Reward-based selection at export.}
The mixed SFT pool applies a reward threshold of 0.5 before message conversion. Dedicated multi-service and safety exports use 0.80, the document-extraction export uses 0.999, and the Harbor-to-message export for ClawMark uses 0.3. These are distinct export configurations. The mixed-pool message converter applies the structural and length checks below without imposing an additional reward threshold.

\paragraph{Message conversion and structural checks.}
Execution records are converted into user, assistant, and tool messages, with assistant tool calls serialized as structured fields. A standardized system message is prepended. Records yielding no usable user--assistant interaction are removed. The converter also discards sequences with fewer than five message entries or more than 20 tool calls. Message counts include system, user, assistant, and tool entries; the minimum-length condition therefore counts messages, rather than dialogue rounds.

Each tool observation longer than 1,000 characters is shortened to its first 1,000 characters followed by \texttt{...[truncated]}. After this preprocessing, the converter sums the characters in message content and JSON-serialized tool-call fields, including the truncation marker. It discards an entire example if this total exceeds 40,000 characters. Tool-observation shortening and whole-example rejection are separate operations. The trainer subsequently applies its configured token-length limit.

\paragraph{Conversion replay.}
Replaying the converter on the 1,809-record mixed SFT candidate snapshot yields 937 demonstrations. The reported rejection categories account for the other 872 records: 494 yield no usable user--assistant interaction, 142 have fewer than five messages, 140 exceed 20 tool calls, and 96 exceed the 40,000-character budget. These counts describe the converter's decisions on this candidate snapshot. The output has mean recorded reward 0.862, mean message count 13.8, and mean tool-call count 6.8.

\paragraph{Relation to data selection and supervision.}
GALA selects representative samples through clustering and geometric distances to reduce redundancy in LLM self-training \citep{xie2026gala}. Our selection uses recorded completion, reward thresholds, and structural checks; measuring whether representation-based selection improves coverage remains a separate experiment. SoFT studies soft targets that balance learning from demonstrations with preserving the base distribution \citep{jing2026soft}. Our SFT objective in Equation~\ref{eq:sft} uses the observed assistant tokens as targets. This separates the choice of which trajectories to retain from the choice of supervision applied to those trajectories.

\section{Published Baseline Sources and Protocols}
\label{app:published-baselines}

\paragraph{Claw-Eval.}
The published reference rows in Table~\ref{tab:claweval-matched} reproduce the Score, Pass@3, and Pass$^3$ columns from Table 3 of \citet{claweval2026} (v3). Its Overall covers the 161 General and 38 Multi-turn tasks, totaling 199; the 101 Multimodal tasks are reported separately in the source's Table 4. Published values are transcribed as reported rather than reconstructed from rounded group scores. The paper runs three trials with temperature zero and no injected service errors, enabling extended thinking where available. Gemini 3 Flash \citep{gemini3flashcard2025} judges open-ended General criteria; Claude Opus 4.6 \citep{claudeopus46card2026} at temperature 0.7 provides the Multi-turn simulator and judge. Our task counts alone do not establish alignment with these settings. The local 8B base and SFT results use the same evaluation configuration, with three trials per task and the same per-trial pass threshold of 0.75. Scores and changes are rounded separately to one decimal place.

\paragraph{ClawMark.}
Table~\ref{tab:clawmark-main} includes seven reference models from the official ClawMark leaderboard \citep{clawmarkleaderboard2026}. Website reference scores use OpenClaw \citep{openclaw2026} Run~1 over the same 100 tasks \citep{clawmarkruns2026}. We use the explicitly named Run~1 field for every model and task and report the official aggregate on a 0--100 scale to one decimal place. This reference protocol uses one trial per task. Our base and SFT results are shown separately, with 100 tasks per model.

\paragraph{Reference-model documentation.}
The reference systems in Tables~\ref{tab:claweval-matched} and~\ref{tab:clawmark-main} are documented in the provider releases for GPT-5.4 \citep{gpt54card2026}, GPT-5.4 mini \citep{gpt54mini2026}, Claude Opus 4.6 and Sonnet 4.6 \citep{claudeopus46card2026,claudesonnet46card2026}, Gemini 3.1 Pro and Gemini 3 Flash \citep{gemini31procard2026,gemini3flashcard2025}, MiMo-V2-Pro and MiMo-V2-Omni \citep{mimov2pro2026,mimov2omni2026}, MiniMax M2.7 \citep{minimaxm27card2026}, and Qwen3.6 Plus \citep{qwen36plus2026}. GLM-5-Turbo has official model documentation \citep{glm5turbodoc}; GLM-5V-Turbo and DeepSeek-V3.2 have technical reports \citep{hong2026glm,liu2025deepseek}. These references identify the models; the benchmark scores and protocols come from the Claw-Eval paper and ClawMark records cited above.

\paragraph{Model scale.}
The official model cards report 120B total / 12B active parameters for Nemotron 3 Super \citep{nemotron3supercard2026}, 1T / 32B for Kimi K2.5 \citep{kimik25card2026}, and 397B / 17B for Qwen3.5-397B-A17B \citep{qwen35card2026}. These are reported MoE parameter counts, not measured inference costs. We do not infer latency or cost savings from total-parameter ratios.

\end{document}